\def\year{2021}\relax
\documentclass[letterpaper]{article} 
\usepackage{aaai21}  
\usepackage{times}  
\usepackage{helvet} 
\usepackage{courier}  
\usepackage[hyphens]{url}  
\usepackage{graphicx} 
\usepackage{natbib}  
\usepackage{caption} 
\usepackage{subfigure}
\usepackage{amsfonts}
\usepackage{caption}
\usepackage{graphicx}
\usepackage{appendix}
\usepackage{float}
\usepackage{amsmath}
\usepackage{enumitem}
\usepackage[export]{adjustbox}
\usepackage{booktabs}
\usepackage[colorinlistoftodos]{todonotes}
\usepackage{multirow}

\nocopyright
\title{Winning Lottery Tickets in Deep Generative Models}
\author {
    Anonymous \\
}

\title{Winning Lottery Tickets in Deep Generative Models}
\author {

        Neha Mukund Kalibhat,\textsuperscript{\rm 1}
        Yogesh Balaji, \textsuperscript{\rm 1}
        Soheil Feizi \textsuperscript{\rm 1} \\
}
\affiliations {
    \textsuperscript{\rm 1} Department of Computer Science, University of Maryland - College Park \\
    \{nehamk,yogesh,sfeizi\}@cs.umd.edu
}

\begin{document}

\maketitle

\begin{abstract}
The lottery ticket hypothesis suggests that sparse, sub-networks of a given neural network, if initialized properly, can be trained to reach comparable or even better performance to that of the original network. Prior works in lottery tickets have primarily focused on the supervised learning setup, with several papers proposing effective ways of finding \textit{winning tickets} in classification problems. In this paper, we confirm the existence of winning tickets in deep generative models such as GANs and VAEs. We show that the popular iterative magnitude pruning approach (with late resetting) can be used with generative losses to find the winning tickets. This approach effectively yields tickets with sparsity up to 99\% for AutoEncoders, 93\% for VAEs and 89\% for GANs on CIFAR and Celeb-A datasets. We also demonstrate the transferability of winning tickets across different generative models (GANs and VAEs) sharing the same architecture, suggesting that winning tickets have inductive biases that could help train a wide range of deep generative models. Furthermore, we show the practical benefits of lottery tickets in generative models by detecting tickets at very early stages in training called \textit{early-bird tickets}. Through early-bird tickets, we can achieve up to 88\% reduction in floating-point operations (FLOPs) and 54\% reduction in training time, making it possible to train large-scale generative models over tight resource constraints. These results out-perform existing early pruning methods like SNIP \cite{lee2018snip} and GraSP \cite{Wang2020Picking}. Our findings shed light towards existence of proper network initializations that could improve convergence and stability of generative models. 
\end{abstract}

\section{Introduction}
The lottery ticket hypothesis \cite{frankle2018lottery}, suggests that there exist sparse sub-networks in over-parameterized neural networks that can be trained to achieve similar or better accuracy than the original network, under the same parameter initialization. These sub-networks and their associated initializations form \textit{winning tickets}. In addition to saving memory, winning tickets have also been shown to achieve improved performance \cite{frankle2018lottery}.

Evidence of the existence of winning tickets has been shown successfully on Visual Recognition tasks (on various CNN-based architectures such as VGG and ResNet)~\cite{Morcos2019OneTT}, Reinforcement Learning and Natural Language Processing tasks \cite{Yu2020PlayingTL}. While research in finding lottery tickets has primarily focused on the classification problem, to the best of our knowledge, no prior work exists on understanding the lottery ticket hypothesis in deep generative models. This will be the focus of our work.

In particular, we investigate if winning tickets exist in two popular families of deep generative models --- Variational AutoEncoders (VAEs) \cite{Kingma2014AutoEncodingVB} and Generative Adversarial Networks (GAN) \cite{Goodfellow2014GenerativeAN}. VAEs are relatively easier to train compared to GANs since their optimization involves a single minimization objective (i.e. the evidence lower bound). Training GANs, on the other hand, is notoriously difficult as it involves a \emph{min-max} game between the generator and the critic networks, causing instability in training ~\cite{Goodfellow2014GenerativeAN, arjovsky2017wasserstein}. In both GANs and VAEs, models are in the over-parameterized regime \cite{brock2018large, razavi2019generating}
to improve the quality of reconstructions and sample generations, especially in large-scale datasets such as Celeb-A \cite{liu2015faceattributes} and ImageNet \cite{imagenet_cvpr09}. This results in significant overhead in training time, compute and memory requirements. Thus, it will be useful, in terms of model storage, training time and training stability, to address the fundamental issue of over-parameterization in deep generative models.

\begin{figure*}[h]
\centering
\includegraphics[width=0.85\textwidth]{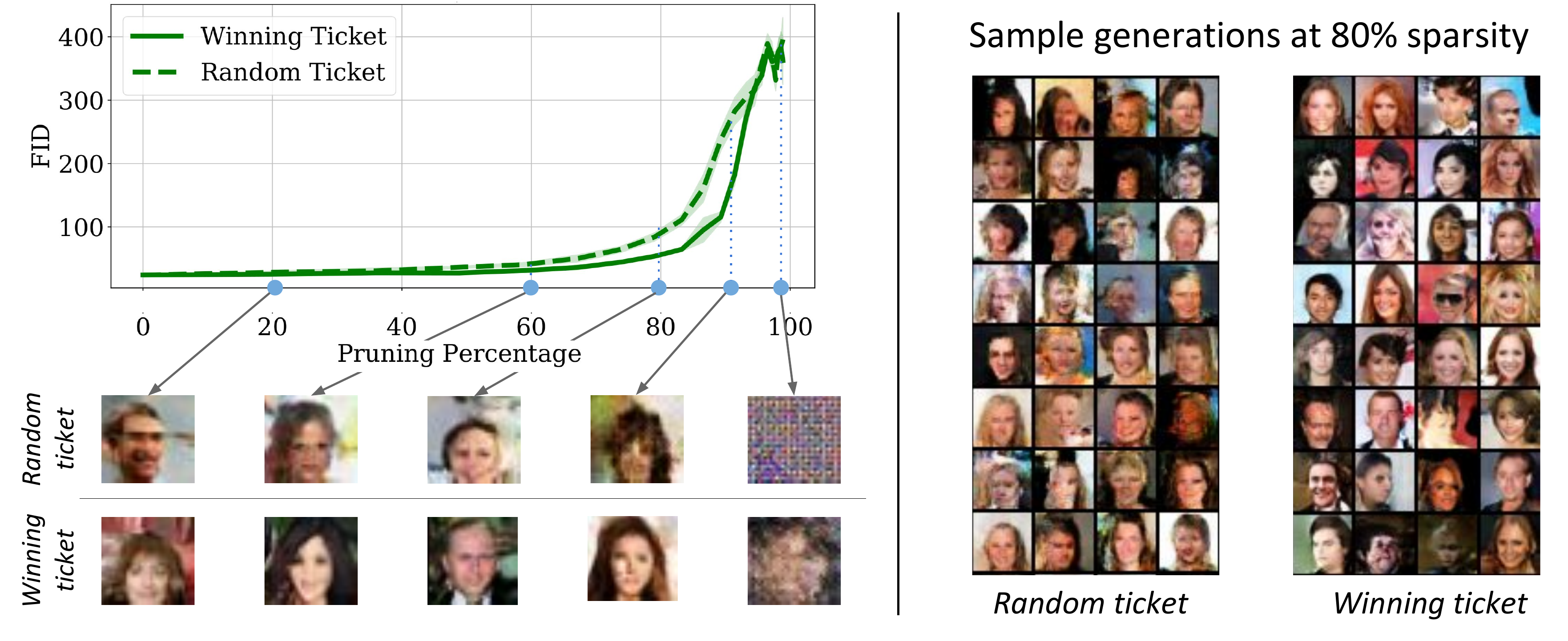}
\caption{\textbf{Lottery Ticket Hypothesis in Generative Models.} The panel on the left shows FID scores of winning tickets and random tickets on a DCGAN model trained on Celeb-A. Winning tickets clearly outperform random tickets at higher pruning regimes. The improved performance of winning tickets is also evident from qualitative results where we find winning tickets generate better quality samples at all sparsity levels. The panel on the right shows sample generations at $80 \%$ sparsity. }
\label{celeba_images} 
\end{figure*}

Another motivation is to see if winning tickets can be {\it transferred} across different generative models. In particular, the generator network in GANs and the decoder network in VAEs fundamentally perform the same task, i.e. transforming the input vectors in the latent space to realistic images. This hints that winning tickets (sub-networks and their initializations) on one generative model (e.g. VAE) might be also a successful ticket on another generative model (e.g. GAN) of the same architecture, although they are based on completely different loss functions. If this is indeed the case, this would imply that a wide range of generative models with different loss functions can share similar network structures and initializations, showing generalizability and scalability of winning tickets. Therefore, to verify transferability of winning tickets, we train the generator of a GAN using the winning ticket obtained from the decoder of VAE, and vice versa, while keeping the other components of both models unpruned. 

The process of finding winning tickets uses a technique called \textit{Iterative Magnitude Pruning}, which involves alternating between network pruning and network re-training steps, while gradually pruning the model. At each iteration of this process, we obtain a sparse sub-network along with its parameter initializations, both of which constitute a \emph{ticket}. We also observe that \textit{late rewinding} \cite{frankle2019linear} is a favorable approach in generative models. Therefore, for each model, we compare the performance of (1) winning ticket, and (2) randomly-initialized ticket. If the randomly initialized sub-network performs significantly worse, a \emph{winning ticket} is obtained. AutoEncoders and VAEs consist of two components: an encoder and a decoder network, while GANs have a generator and a discriminator. In each case, we perform iterative pruning experiments on either jointly pruning both networks of the model or by pruning each network separately while leaving the other unpruned. 

In contrast to classification tasks where there is a well-defined evaluation metric (i.e. classification accuracy) for assessing the model performance, we do not have such a metric in deep generative modeling. On image-based datasets, Fréchet Inception Distance (FID) \cite{heusel2017gans} is one popular metric used in evaluating deep generative models. Figure \ref{celeba_images} shows winning and random tickets evaluated on FID and also illustrates the differences in generated images. We have also used metrics such as the reconstruction loss (in AutoEncoders), the discriminator loss (in GANs) and the downstream classification accuracy (in AutoEncoders and VAEs).

The existence of winning tickets indicates that generative models can be trained under limited memory and resource constraints. However, finding these tickets requires multiple rounds of training, and each training cycle can last for weeks for large models such as BigGAN(\cite{brock2018large}). Moreover, the sparse sub-networks found using iterative magnitude pruning have individual parameter-level sparsity (as opposed to channel-level sparsity). Hence, minimizing compute is not possible without using specialized hardward \cite{cerebras, nvidia}. Instead, we are interested in finding better pruning strategies that provides compute gains on existing hardware. To this end, we investigate the effectiveness of \textit{early-bird tickets}, which are channel-pruned sub-networks found early in the training \cite{You2020Drawing} in the context of generative models. We also compare the performance of early-bird tickets with other pruning strategies like SNIP \cite{lee2018snip} and GraSP \cite{Wang2020Picking} that prune the network at initialization.


\begin{table*}[h]
\caption{Summary of lottery ticket experiments conducted on various generative models}
\centering
\resizebox{0.85\textwidth}{!}{
 \begin{tabular}{l l|c|c|c|c} 
 \toprule
 & \textbf{\multirow{2}{*}{Network}} & \textbf{Number of} & \textbf{\multirow{2}{*}{Datasets}} & \textbf{Winning Ticket} & \textbf{\multirow{2}{*}{Evaluation Metric}} \\[0.5ex]
 &&\textbf{Parameters}&& \textbf{Sparsity}& \\
 \midrule
 \midrule
 \multirow{3}{*}{\rotatebox{90}{\centering \textit{AE} }} &Linear AutoEncoder & 200K & MNIST & 89.2\% & \multirow{3}{*}{Reconstruction Loss, Test Accuracy} \\

 & \multirow{2}{*}{Conv. AutoEncoder} & \multirow{2}{*}{3M} & CIFAR-10 & 95.6\% &  \\
 & & & Celeb-A & 98.5\% &  \\
\cmidrule{1-6}
 \multirow{3}{*}{\rotatebox{90}{\centering \textit{VAE} }} & VAE \cite{Kingma2014AutoEncodingVB} & 5.6M & \multirow{3}{*}{CIFAR-10, Celeb-A} & 79\% & \multirow{3}{*}{FID, Test Accuracy} \\

 & $\beta$-VAE \cite{Higgins2017betaVAELB} & 5.6M &  & 86.5\% & \\

 & ResNet-VAE \cite{kingma2016improving} & 2.8M &  & 93.1\% &  \\
\cmidrule{1-6}
 \multirow{5}{*}{\rotatebox{90}{\centering \textit{GAN} }} & DCGAN \cite{radford2015unsupervised} & 6.5M & \multirow{3}{*}{CIFAR-10, Celeb-A} & 83.2\% & \multirow{5}{*}{FID, Discriminator Loss} \\

 & SNGAN \cite{miyato2018spectral} & 6.7M &  & 89.2\% & \\

 & ResNet-DCGAN & 2.2M &  & 79\% &  \\
 
 & \multirow{2}{*}{WGAN~\cite{arjovsky2017wasserstein}}  & \multirow{2}{*}{6.5M} & CIFAR-10  & 73.7\% &  \\
 
 &  & & Celeb-A  & 59\% &  \\
 \bottomrule
\end{tabular} 
}

\label{exp_table}
\end{table*}

\begin{figure}[h]
\centering
\includegraphics[width=0.32\textwidth, trim={0.8cm 0.9cm 0.8cm
1.1cm}]{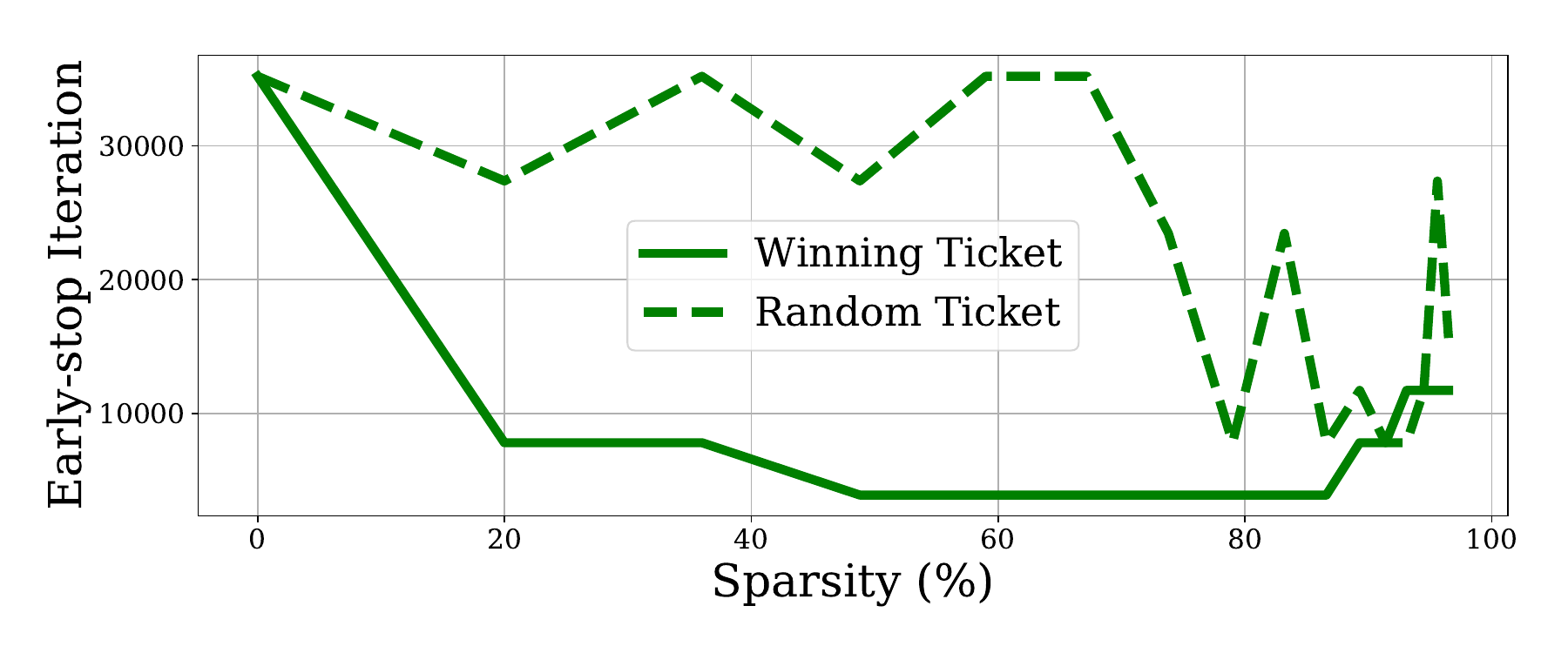}
\caption{\textbf{Convergence of Winning Tickets} The plot shows the early-stopping iteration of lottery tickets. Winning lottery tickets show significantly faster convergence than random tickets. Experiments are performed on $5$ random runs, and the error bars represent $+/-$ standard deviation across runs.}
\label{early} 
\end{figure}

We conduct experiments on several generative models including linear AutoEncoder, convolutional AutoEncoder, VAE, $\beta$-VAE \cite{Higgins2017betaVAELB}, ResNet-VAE \cite{kingma2016improving}, Deep-Convolutional GAN (DCGAN) \cite{radford2015unsupervised}, Spectral Normalization GAN (SNGAN) \cite{miyato2018spectral}, Wasserstein GAN (WGAN) \cite{arjovsky2017wasserstein} and ResNet-GAN \cite{He_2016} on MNIST \cite{lecun2010mnist}, CIFAR-10 \cite{Krizhevsky09learningmultiple} and Celeb-A \cite{liu2015faceattributes} datasets. Table \ref{exp_table} summarizes all our experiments and the winning ticket sparsity achieved for each model. We make the following observations:

\begin{itemize}[leftmargin=*]
    \item \textbf{AutoEncoders: } We find winning tickets of sparsity 89\% on MNIST, 96\% on CIFAR-10 and 99\% on Celeb-A. 
    \item \textbf{VAEs:} We find winning tickets of sparsity 79\% on VAE, 87\% on $\beta$-VAE and 93\% on ResNet-VAE on both datasets.
    \item \textbf{GANs:} We find winning tickets of sparsity 83\% on DCGAN, 89\% on SNGAN and 79\% on ResNet-DCGAN on both datasets. In WGAN we find winning tickets at 73.7\% (CIFAR-10) and 59\% (Celeb-A).
    \item \textbf{Single Component Pruning:} In AutoEncoders and GANs, comparable sparsities of both components is essential for the best model performance. In VAEs, on the other hand, encoder-only pruning preserves performance, implying that VAEs can be trained with very sparse encoders.
    \item \textbf{Late Rewinding:} For very deep generative models, we observe that rewinding weights to an early iteration instead of initialization, is favorable for better stability and performance of winning tickets.
    \item \textbf{Stability:} Winning tickets demonstrate higher stability across multiple runs compared to random tickets.
    \item \textbf{Convergence:} Winning tickets converge significantly faster than the unpruned network and random tickets. This is demonstrated in Figure \ref{early}.
    \item \textbf{Transferability:} Winning tickets transferred from VAEs to GANs perform on par to winning tickets solely found on GANs and vice versa. 
    \item \textbf{Early-Bird Tickets:} Early-Bird tickets reduce the training time by 54\% and FLOPs by 88\%. They outperform other early pruning strategies like SNIP \cite{lee2018snip} and GraSP \cite{Wang2020Picking} in terms of FID, FLOPs and training time. 
\end{itemize}

These results shed some light on existence of proper network initializations and architectures in deep generative models that could improve their computational and statistical properties such as their convergence, stability and storage. In particular, the transferability of winning tickets across seemingly different generative models such as GANs and VAEs suggest that there exists unifying network architectures and initialization strategies across these models. 

\section{Related Work}
{\bf Generative Models.} Two prominent and popular deep generative models are VAEs and GANs. VAEs train a generative model by maximizing an Evidence Lower Bound (ELBO) with an encoder-decoder structure, in which the encoder maps the data samples to a latent space, while the decoder reconstructs the latent representations back to the input space \cite{Kingma2014AutoEncodingVB}. $\beta$-VAE \cite{Higgins2017betaVAELB} proposes a modification to the VAE objective, where an adjustable hyper-parameter $\beta$ balances the reconstruction accuracy and latent representation constraints, leading to improved qualitative performance. On the other hand, GANs \cite{Goodfellow2014GenerativeAN} train a generative model by transforming input samples with known tractable distribution (such as Gaussian) to an unknown data distribution (e.g. images). This transformation function is learnt using an adversarial game between generator and discriminator networks. This \emph{min-max} game results in difficulties in optimization, especially in deep networks. To improve the stability, several techniques such as spectral normalization~\cite{miyato2018spectral}, Wasserstein losses~\cite{arjovsky2017wasserstein}, gradient penalties~\cite{gulrajani2017improved}, self-attention models \cite{zhang2018selfattention}, etc. have been proposed.  

{\bf Generative Model Pruning.} Network pruning and model compression are important topics in machine learning especially in supervised learning setups \cite{Cun90optimalbrain, 10.5555/2987189.2987223, article, li2016pruning}. However, these problems have been relatively less explored in deep generative models. The magnitude pruning approach used in this paper \cite{Han2015LearningBW}, zeroes out weights with small magnitudes, followed by re-training. Other pruning approaches include pruning groups of weights together (i.e. structured sparsity learning) \cite{wen2016learning} , pruning filters and their connecting edges \cite{li2016pruning} and enforcing channel-level sparsity \cite{Liu_2017}. SNIP \cite{lee2018snip} and \cite{Wang2020Picking} are recent approaches that propose pruning at initialization. Knowledge distillation-based approaches~\cite{hinton2015distilling}, in which smaller networks are trained using the distillation loss from a larger teacher network have also been successfully used in compressing GANs \cite{aguinaldo2019compressing,koratana2018lit}.

{\bf Lottery Ticket Hypothesis.} The lottery ticket hypothesis, \cite{frankle2018lottery} proposes Iterative Magnitude Pruning (IMP) to find tickets. In deeper networks, it has been shown that IMP with late-rewinding  \cite{frankle2019linear, Morcos2019OneTT} is more beneficial than rewinding to initialization. While many early pruning strategies \cite{lee2018snip, Wang2020Picking} have been proposed, it has been shown that these methods often fall-short compared to IMP \cite{frankle2020pruning}. It is also shown \cite{zhou2019deconstructing} that signs of the weights are more important than their magnitudes and as long as they remain the same, the sparse model can still train more successfully than with random sign assignment initializations. However, a somehow contradicting observation is made \cite{frankle2020early}, where authors show that deep networks are not robust to random weights while maintaining signs. It has also been shown that winning tickets reflect inductive biases and do not over-fit to particular domains \cite{Desai_2019}. Although the initial work on the hypothesis has focused on supervised image classification tasks, successful results on other tasks have been observed too. Transfer learning tasks \cite{mehta2019sparse}, transformer models in Natural Language Processing and Reinforcement Learning \cite{Yu2020PlayingTL} have successfully uncovered winning tickets. The lottery ticket hypothesis however, has been challenged in \cite{liu2018rethinking} which argues that randomly initialized tickets can match performance to winning tickets if trained with an optimal learning rate and for long enough. 

\section{Methods}

\subsection{Pruning Approach}
\subsubsection{Winning Lottery Tickets}
\label{winning_tickets_section}
\begin{figure}[h]
\subfigure{\includegraphics[width=0.23\textwidth, trim={0.9cm 0.9cm 1cm
0.9cm}, clip]{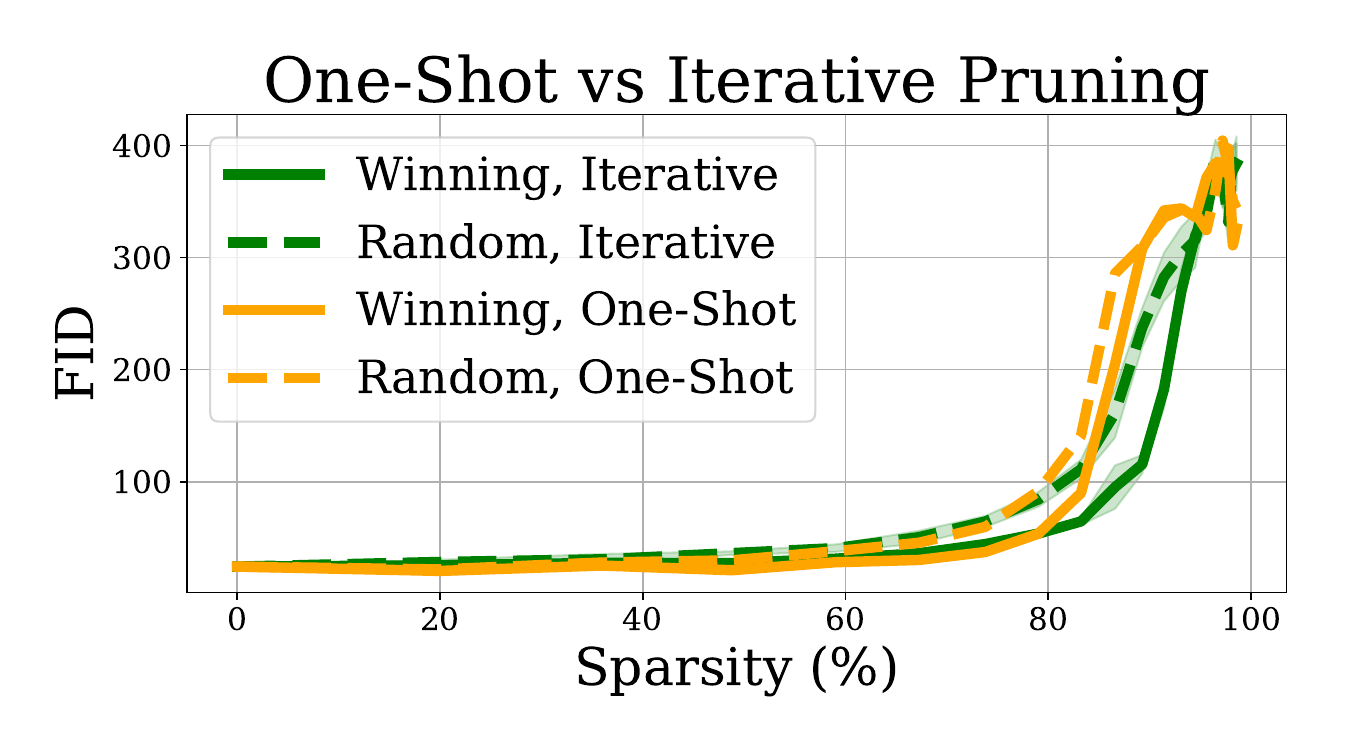}}
\subfigure{\includegraphics[width=0.23\textwidth, trim={0.9cm 0.9cm 1cm
0.9cm}, clip]{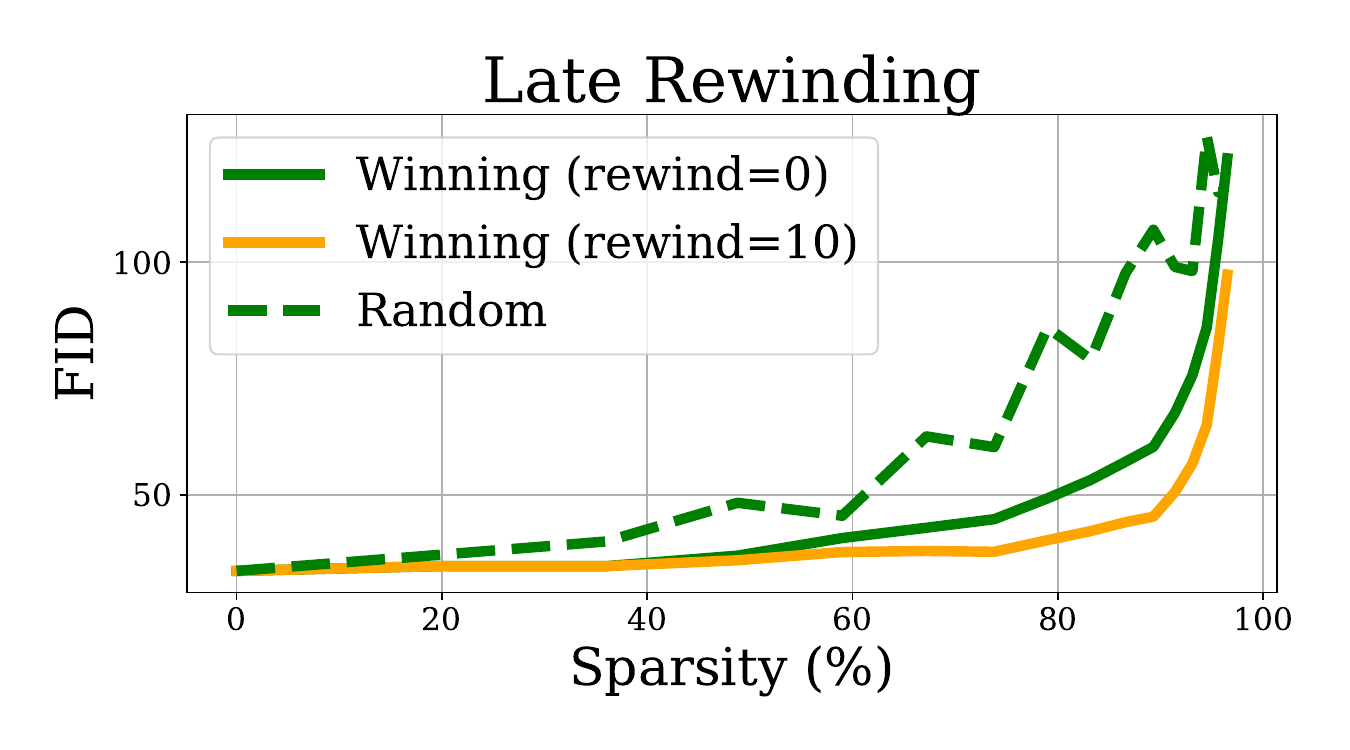}}
\caption{\textbf{Pruning Strategy} The plot on the left shows the comparison of one-shot and iterative pruning on DCGAN trained on CIFAR-10. The plot on the right shows FID of tickets rewound to initialization and to iteration 10. Iterative pruning and late rewinding shows better tickets at high sparsities. Experiments are performed on $5$ random runs, and the error bars represent $+/-$ standard deviation across runs.}
\label{one_shot_late} 
\end{figure}

One-shot pruning and Iterative Magnitude Pruning are standard approaches to find winning lottery tickets. In one-shot pruning, the lower $p\%$ weights of the trained network with initialization $\theta_0$ (i.e. parameters with the smallest magnitudes) are pruned and the remaining weights are re-initialized to $\theta_0$ and re-trained. In practice however, pruning a large fraction of weights through one-shot pruning might null the weights that are actually important to the model leading to a significant drop in the performance \cite{Morcos2019OneTT}. In Figure \ref{one_shot_late}, we confirm that this phenomenon applies to generative models as well.

 We instead employ \textit{Iterative Magnitude Pruning} (IMP). Here, we take a trained model initialized with $\theta_0$ and we choose a small pruning percentage $p$. At the first pruning cycle, we one-shot prune $p\%$ of network, generating a mask $m_1$ and re-train the network using $(\theta_0, m_1)$. In the next pruning cycle, $p\%$ of the remaining weights from the previous cycle are one-shot pruned (generating $m_2$) and re-trained with $(\theta_0, m_2)$. We repeat this process of pruning and re-training with masks for $n$ rounds. In this paper, we use a global pruning scheme across all experiments; i.e. weights of all the layers of the network are pooled together and pruned. In all our experiments in this paper, $p = 20\%$ and $n = 20$, i.e. we run 20 rounds of iterative magnitude pruning where we prune 20\% of the network at each iteration. Since $p$ is not too large, it properly scans pruning fractions between 20\% to 98.84\%.

It has been shown that rewinding the network to the weights at training iteration i, $\theta_i$ (where $i \ll N$, N being the total number of training iterations) is better than rewinding to $\theta_0$ \cite{frankle2019linear} as deep neural networks become more stable to noise after a few iterations of training. We confirm this behavior for generative models in Figure \ref{one_shot_late}. Therefore, unless specified otherwise, we apply late-rewinding to all our experiments.

\begin{figure*}[h]
\subfigure{\includegraphics[width=0.24\textwidth, trim={0.9cm 0.9cm 1cm 0.9cm}, clip]{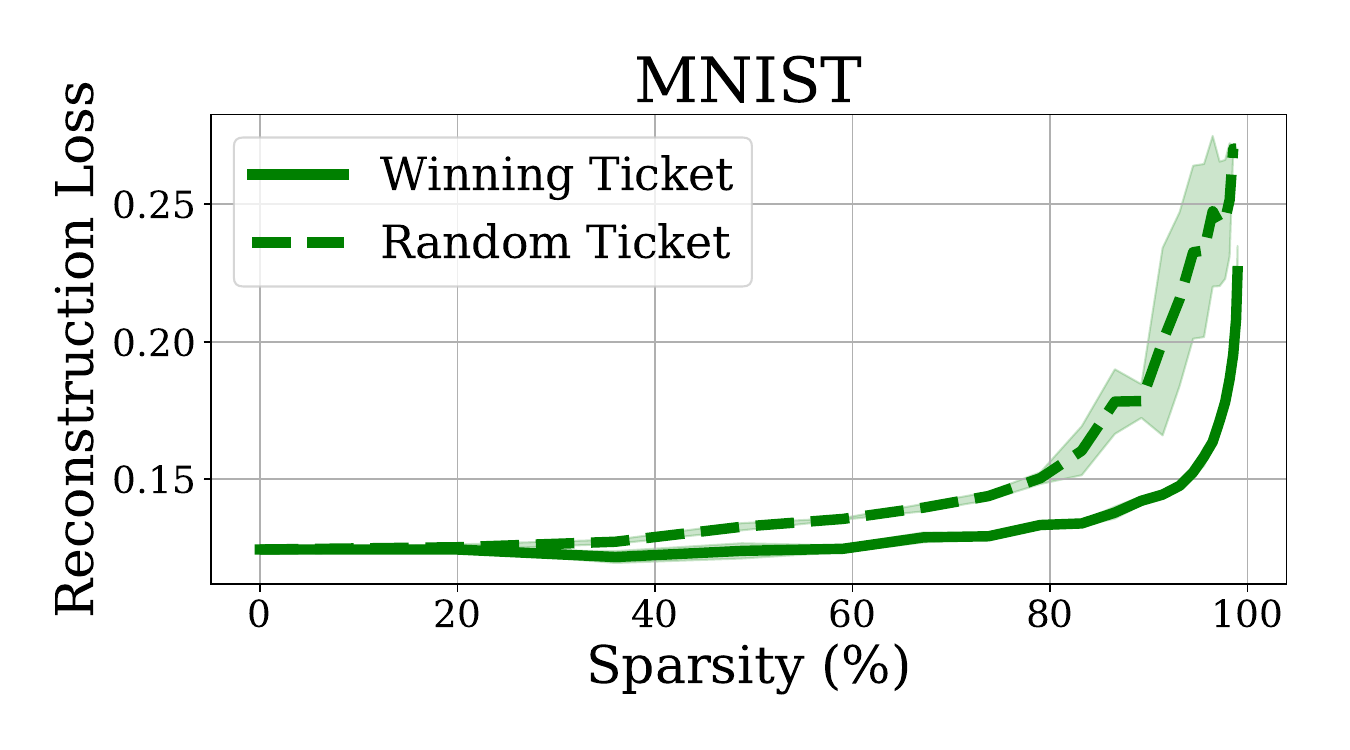}}
\subfigure{\includegraphics[width=0.24\textwidth, trim={0.9cm 0.9cm 1cm 0.9cm}, clip]{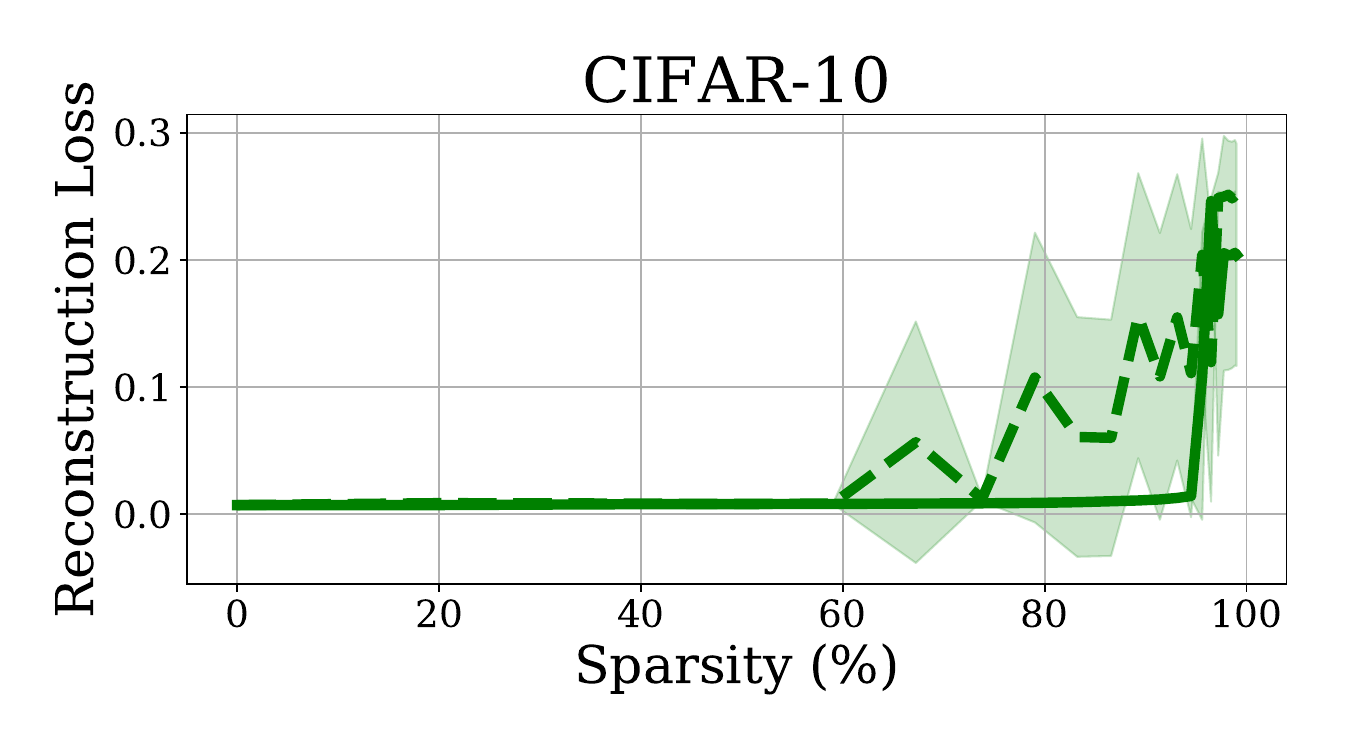}}
\subfigure{\includegraphics[width=0.24\textwidth, trim={0.9cm 0.9cm 1cm 0.9cm}, clip]{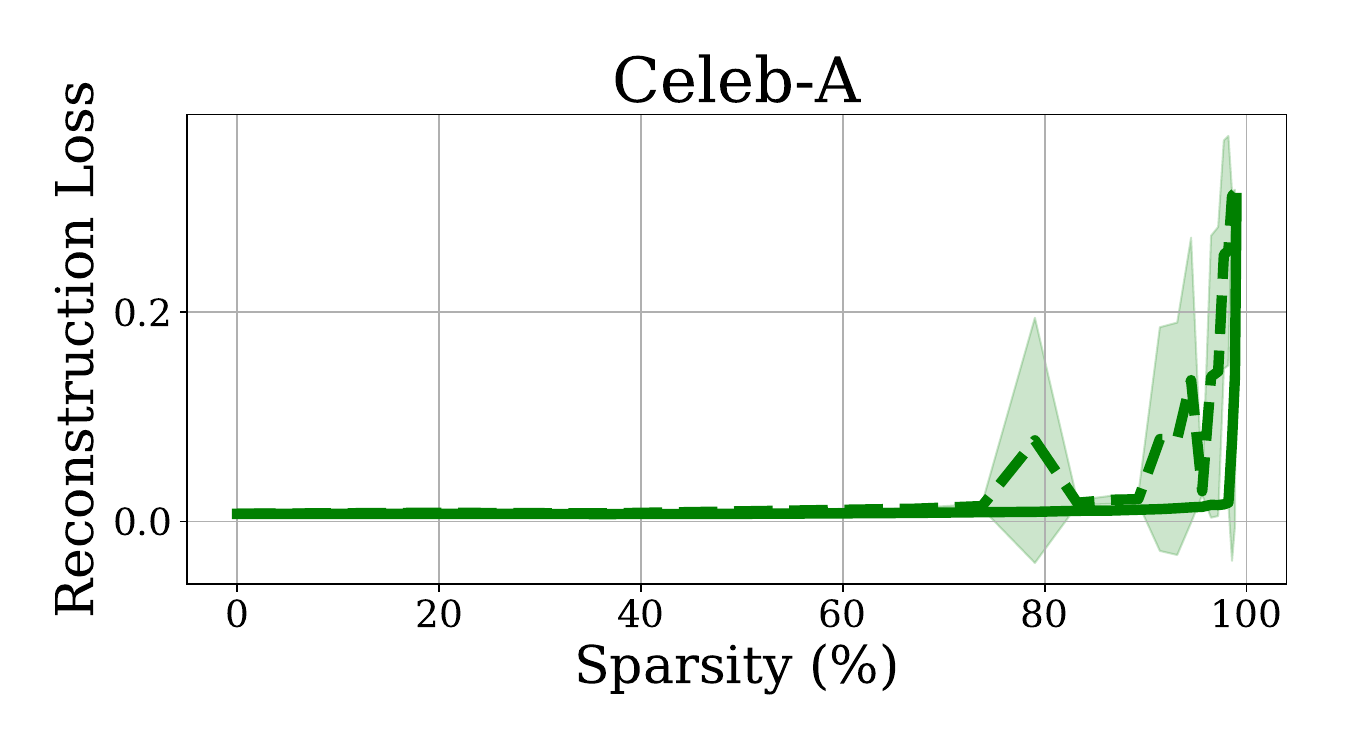}}
\subfigure{\includegraphics[width=0.24\textwidth, trim={0.9cm 0.9cm 1cm 0.9cm}, clip]{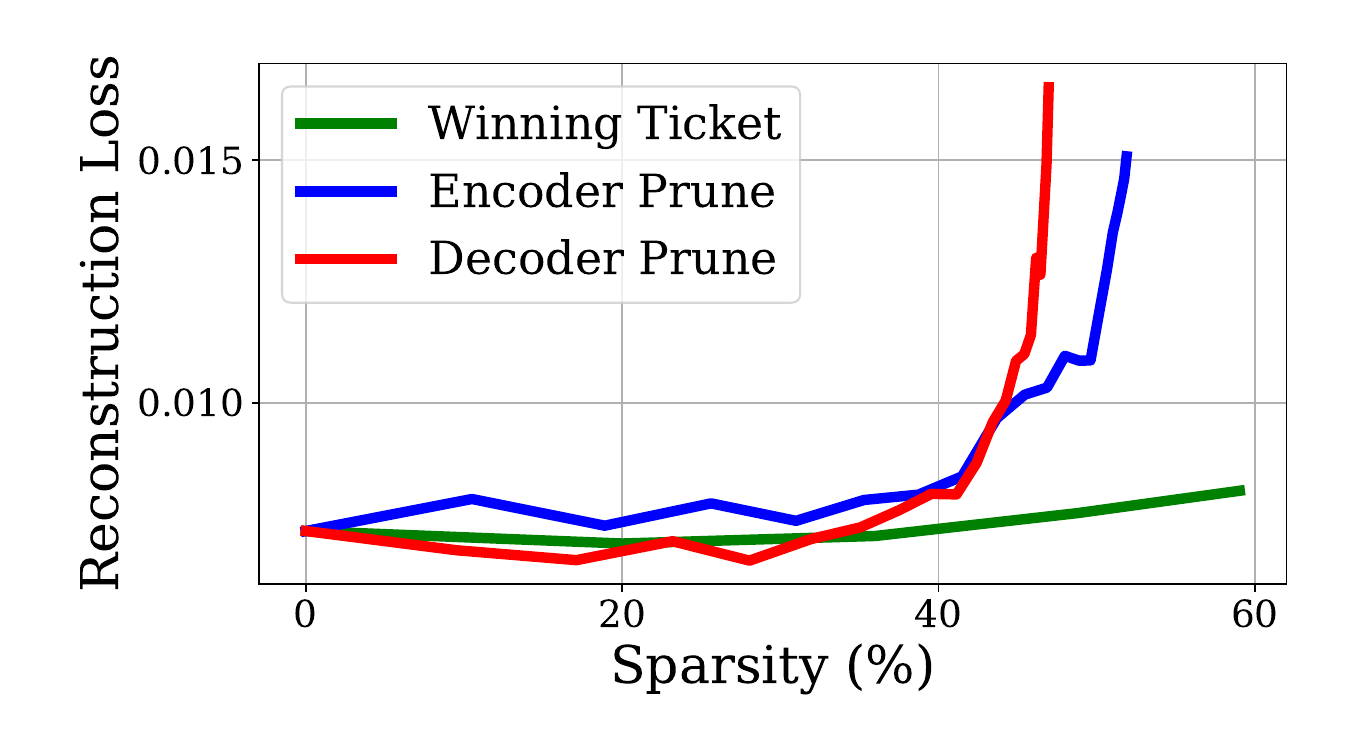}}
\caption{\textbf{AutoEncoder experiments.} We plot reconstruction losses of tickets at different levels of sparsity on MNIST, CIFAR-10 and Celeb-A. In all datasets, performance of winning tickets is consistently better than random tickets. Each experiment is performed on $5$ random runs, and the error bars represent $+/-$ standard deviation across runs.}
\label{auto_plots}
\end{figure*}

\begin{figure*}[h]
\centering
\subfigure{\includegraphics[width=0.3\textwidth, trim={0.9cm 0.9cm 1cm
0.9cm}, clip]{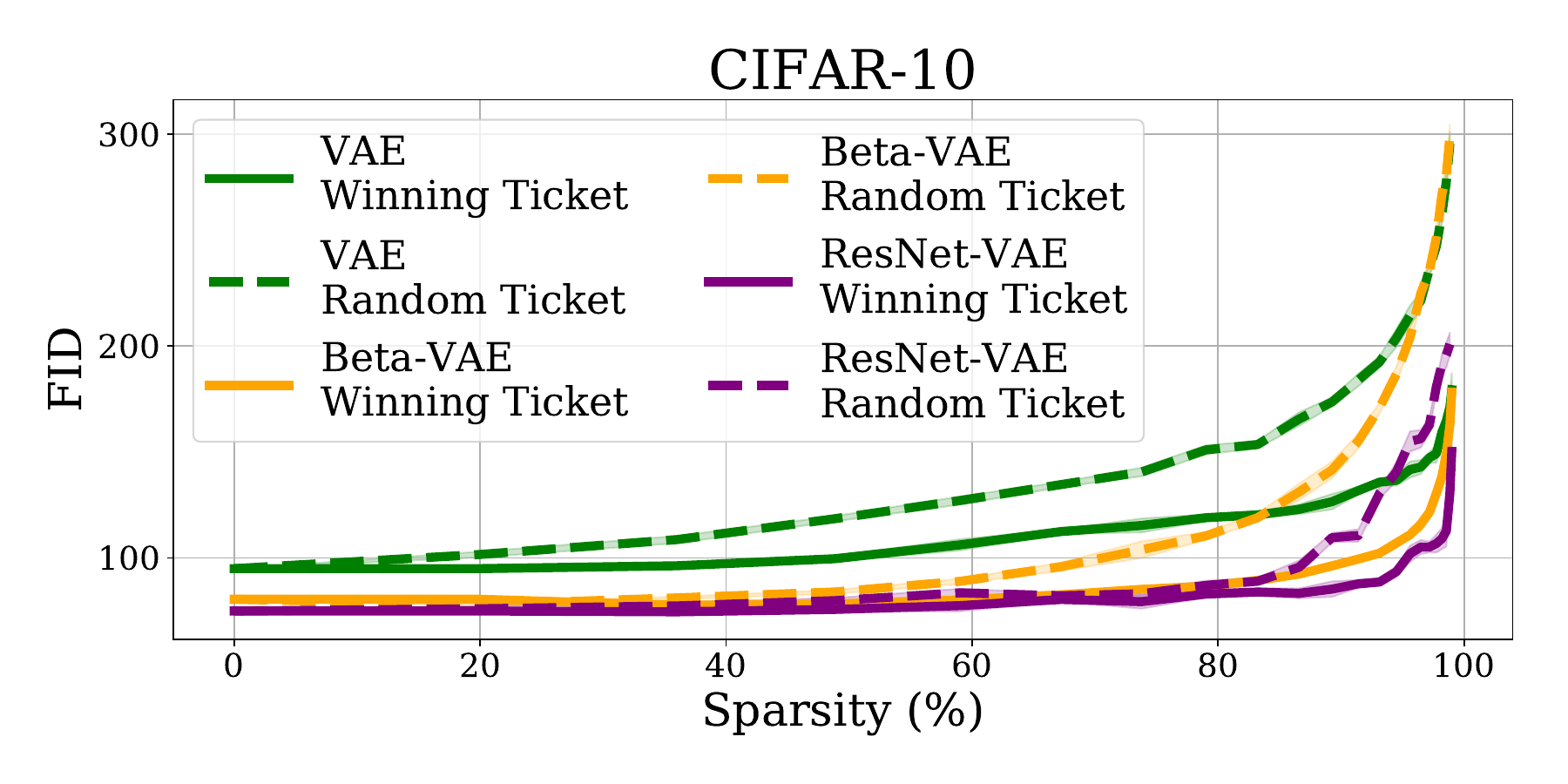}}
\subfigure{\includegraphics[width=0.3\textwidth, trim={0.9cm 0.9cm 1cm
0.9cm}, clip]{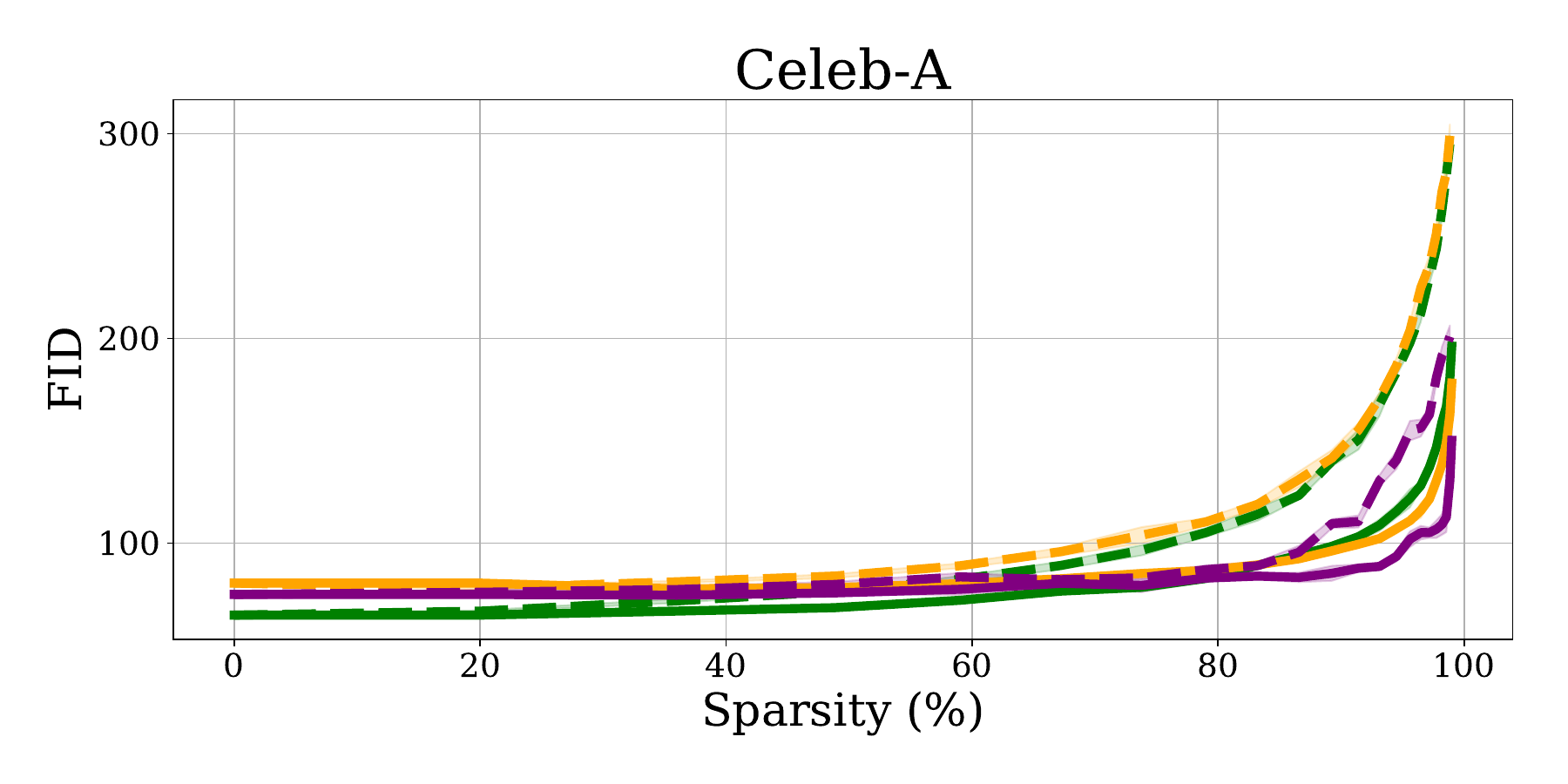}}
\subfigure{\includegraphics[width=0.3\textwidth, trim={0.9cm 0.9cm 1cm 0.9cm}, clip]{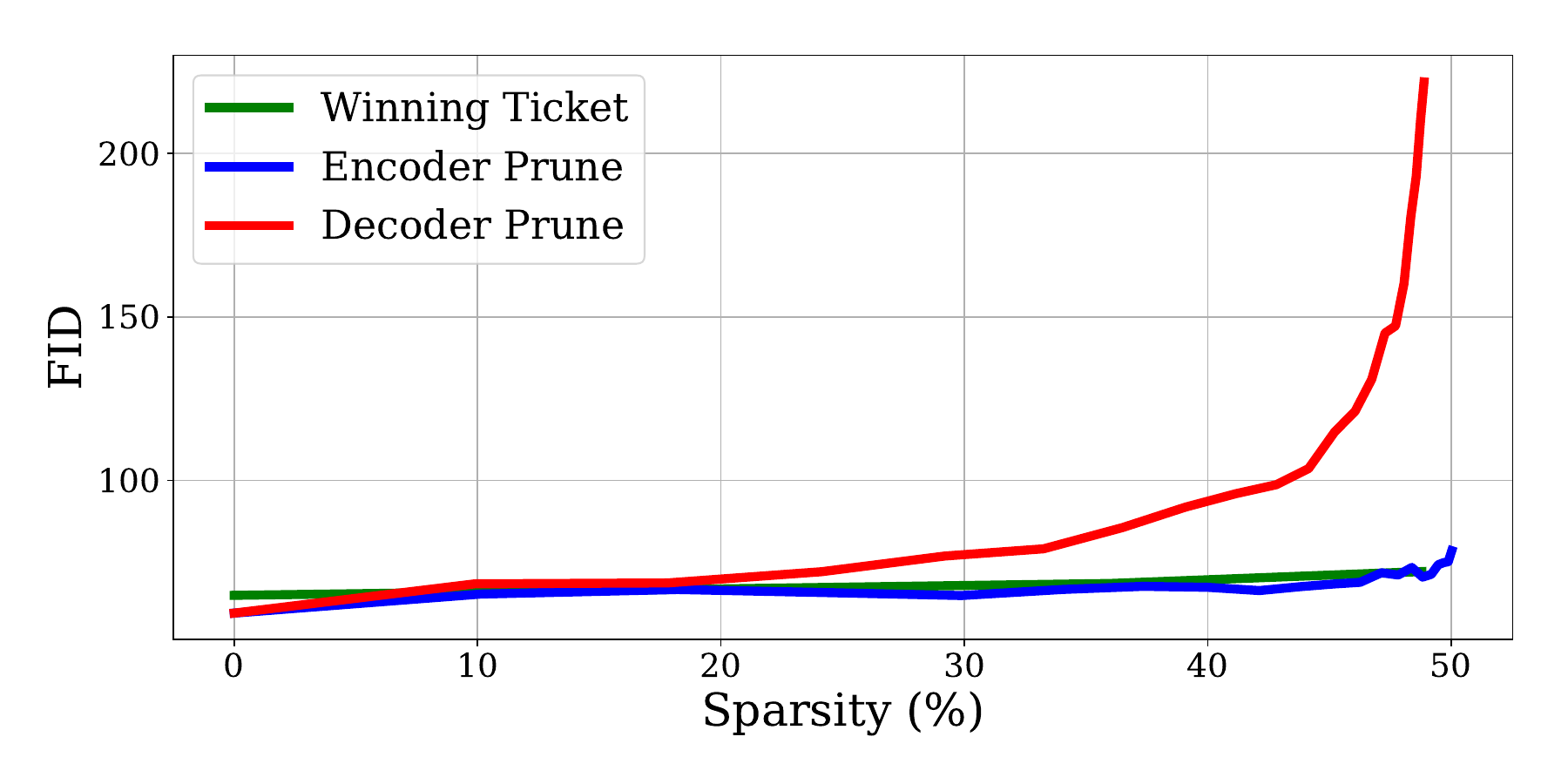}}
\caption{\textbf{VAE experiments.} We plot the FID scores of tickets on CIFAR-10 and Celeb-A on three models: VAE, $\beta$-VAE, and ResNet-VAE. Winning tickets outperform random tickets on all models. Experiments are performed on $5$ random runs, and the error bars represent $+/-$ standard deviation across runs.}
\label{vae_plots}
\end{figure*}

\subsubsection{Early-bird Tickets}
Winning tickets found using IMP is an expensive process involving multiple training cycles for finding the optimal ticket. This is impractical in real-world applications, especially for generative models. In the recent work of \cite{You2020Drawing}, it has been shown that lottery tickets emerge at very early stages of training, also termed as \textit{early-bird tickets} (EB-tickets). EB-tickets reduce the overhead of iterative pruning by finding the ticket early, without having to train models to convergence. Once identified, training can be continued on just the EB-tickets towards convergence. 

The process of finding EB-tickets involves using channel pruning, in which a fraction of batch-normalization channels of the network are pruned \cite{You2020Drawing}. While magnitude pruning removes individual parameters from a neural net, channel pruning effectively removes the entire channel, since setting the affine batch normalization parameters has the same effect of removing the corresponding channel it connects to. This pruning technique results in a shallower network and yields savings in memory and compute without any additional hardware requirements. At every training iteration $i$, we perform channel pruning to get a mask $m_i$. We then compute the \textit{mask distance}, defined as the Hamming distance between $m_i$ and $m_{i-1}$ (look-back can be over multiple iterations). When the mask distance is less than an upper bound $\delta$, an EB-ticket is found. We then compress the network channels and continue training the EB-ticket. In our experiments, we look-back 5 iterations and fix $\delta$ as 0.1. These hyper-parameters generally help us find stable EB-tickets very early in training at epoch 4 to 6. We also perform mixed-precision training on EB-tickets, where the floating-point precision of parameters and their gradients are reduced from 32-bit to 16-bit or 8-bit depending on their sizes. 


\subsection{Evaluation}
\subsubsection{Winning Lottery Tickets}
Tickets discovered in the iterative pruning process need to be tested to verify if the cause of their good performance is their initialization. For such an evaluation, we measure the performance of winning tickets when they are randomly initialized called, \textit{random tickets}. 

For generative models, we use the following metrics: (1) For AutoEncoders, we use the reconstruction loss after the model has converged. We also measure downstream classification accuracy, in which we pass the reconstructed images to a ResNet-18 model trained on CIFAR-10 and Celeb-A, and calculate the test accuracy on the reconstructed samples. (2) In VAEs, we use FID to assess the quality of generated samples. The FID score~\cite{heusel2017gans} calculates the Fréchet Inception distance between the feature distributions (as given by a pre-trained Inception network) of real and generated samples. These feature distributions are approximated using a Gaussian distribution. Then, for two distributions $p_r$, $p_g$ with feature means and co-variances $(\mu_r, \Sigma_r)$ and $(\mu_g, \Sigma_g)$ we calculate
\begin{align*}
    FID(p_r, p_g) = \|\mu_r - \mu_g\|^2 + Tr(\Sigma_r + \Sigma_g) - 2(\Sigma_r \Sigma_g)^{\frac{1}{2}}
\end{align*}
A small change in FID may show little or no change in the quality of the generated samples. However, a large change in FID (e.g. a change $\gg 10$) translates to a perceptible difference in the quality of generated samples. We also use the downstream classification test accuracy as an additional evaluation metric since FID scores in VAEs are often high due to the blurriness of generated images. (3) In GANs, we evaluate winning tickets using FID and the discriminator loss. In addition to these quantitative metrics, we also evaluate images generated by the winning tickets qualitatively. Winning tickets should preserve and generate higher quality images than random tickets. Table \ref{exp_table} summarizes the evaluation metrics used for each generative model. 

\subsubsection{Early-bird Tickets}
We evaluate early-bird tickets in generative models, similar to winning lottery tickets using FID as discussed in the previous section. In addition to FID, we measure the resource usage both in terms of: (1) Floating-point operations (FLOPs) and (2) Training Time. FLOPs are measured by cumulatively adding the floating-point operations of forward and backward propagations of convolutional, linear and batch-normalization layers over the entire training cycle, including pruning operations, giving us an exact measure of the total processor usage.
In addition to FLOPs, we also measure the total training time in seconds.

\begin{figure*}[h]
\centering
\subfigure{\includegraphics[width=0.195\textwidth, trim={0.9cm 0.9cm 1cm 0.9cm}, clip]{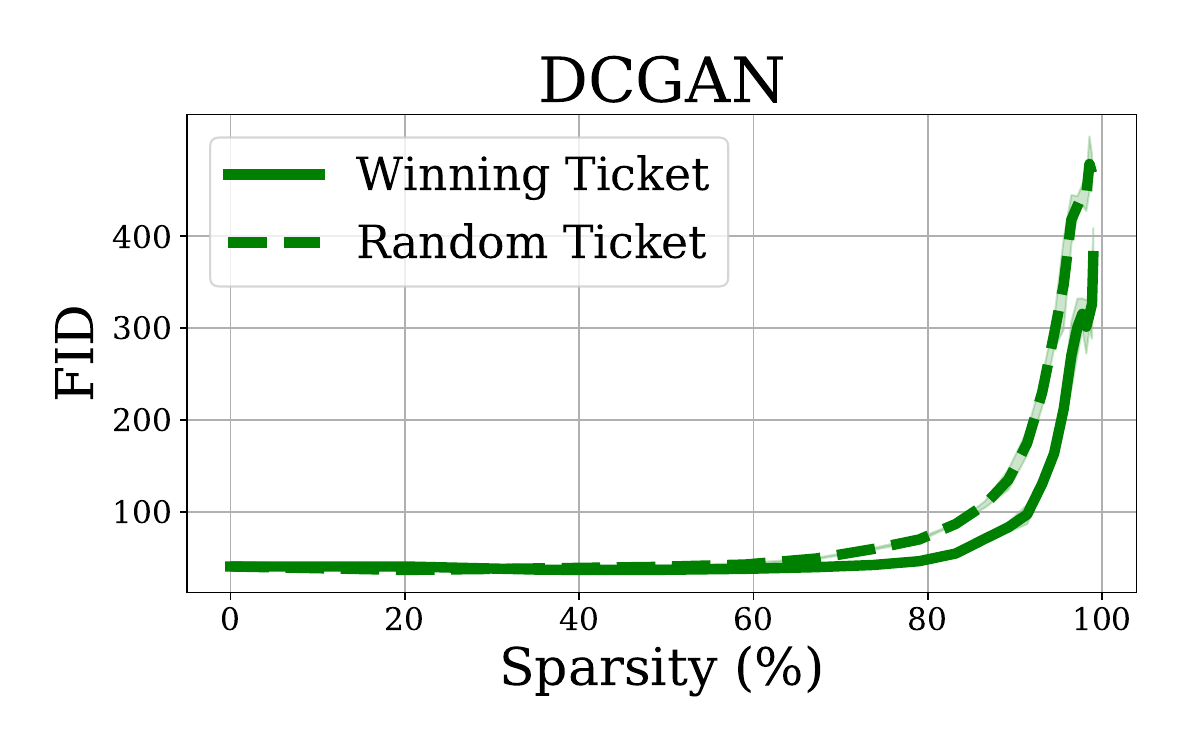}}
\subfigure{\includegraphics[width=0.195\textwidth, trim={0.9cm 0.9cm 1cm 0.9cm}, clip]{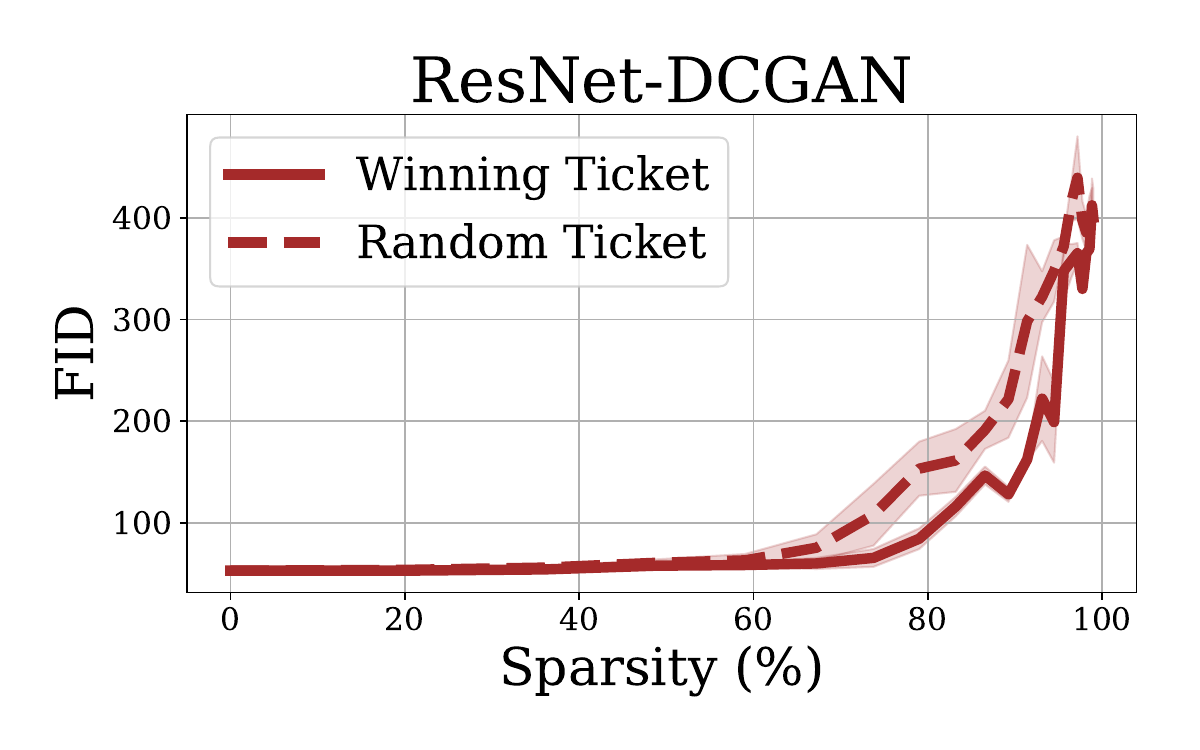}}
\subfigure{\includegraphics[width=0.195\textwidth, trim={0.9cm 0.9cm 1cm 0.9cm}, clip]{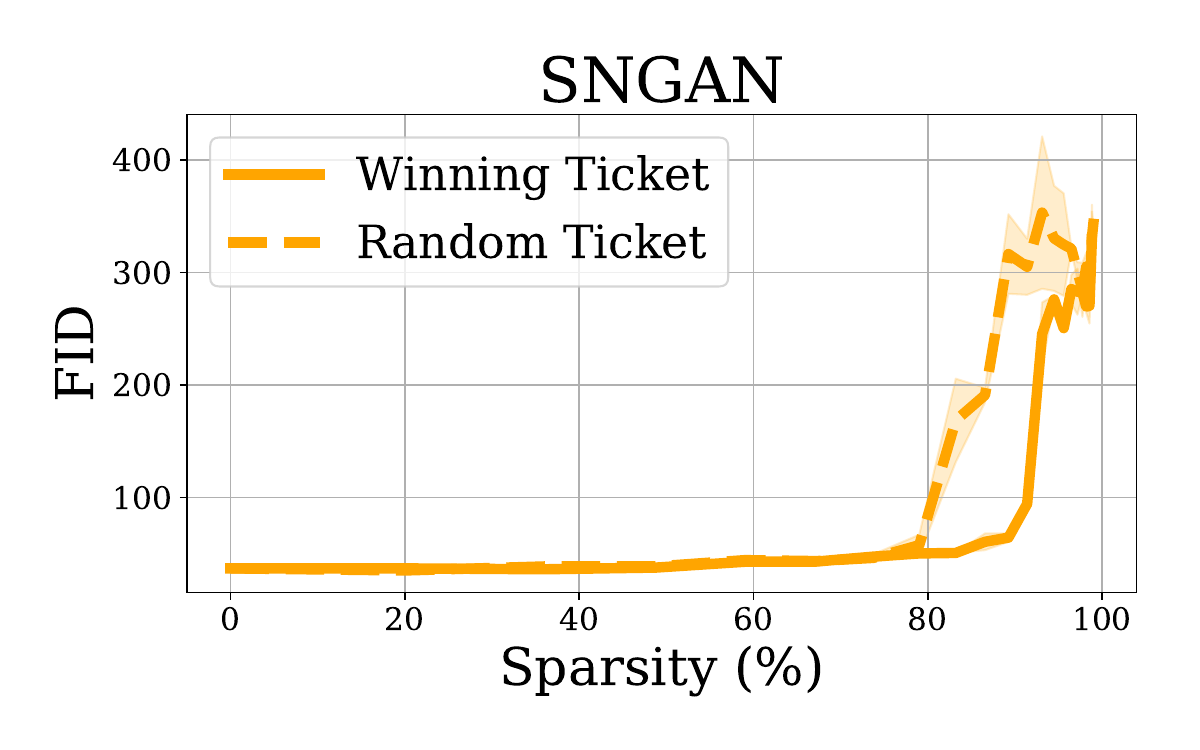}}
\subfigure{\includegraphics[width=0.195\textwidth, trim={0.9cm 0.9cm 1cm 0.9cm}, clip]{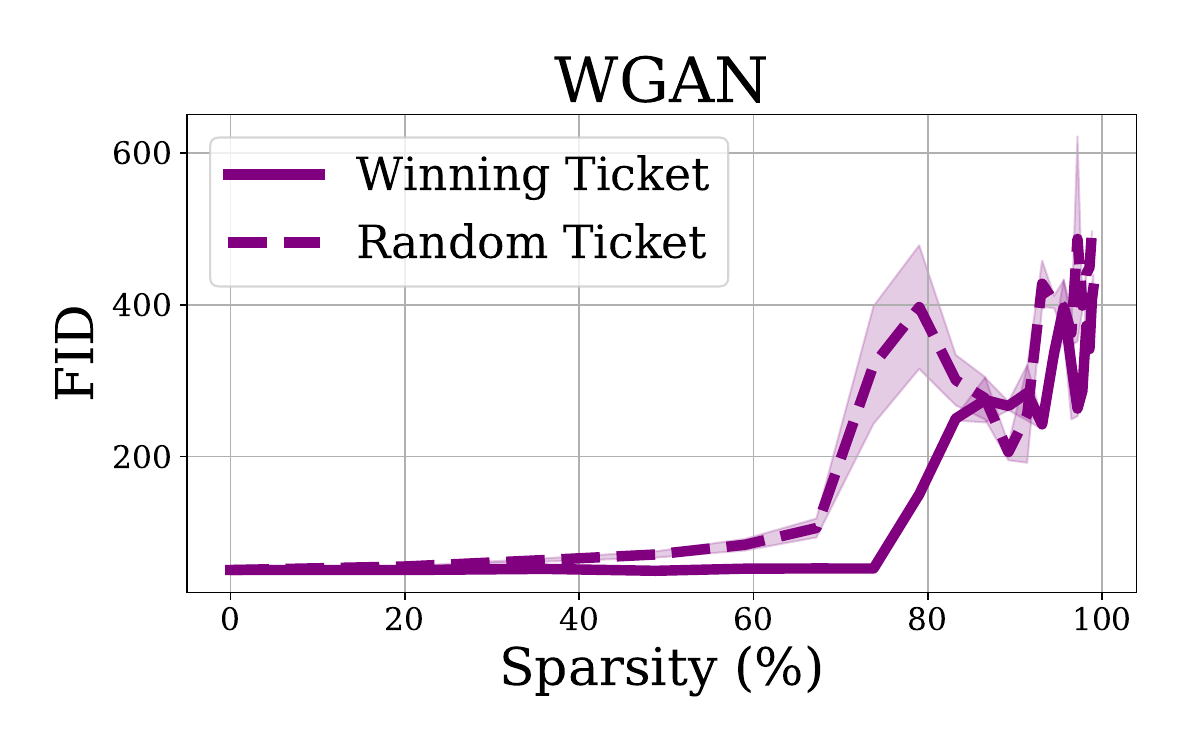}}
\subfigure{\includegraphics[width=0.195\textwidth, trim={0.9cm 0.9cm 1cm 0.9cm}, clip]{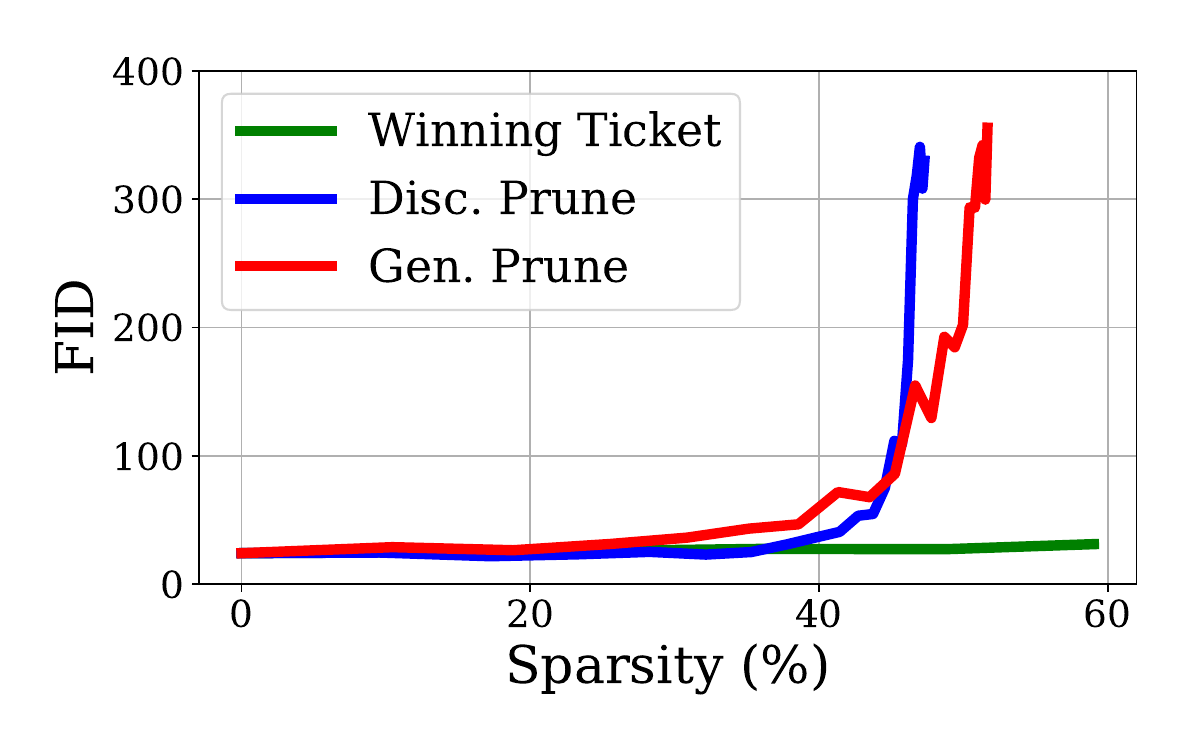}}
\caption{\textbf{GAN experiments. } We plot the FID scores of tickets of $4$ GAN models trained on CIFAR-10 dataset: DCGAN, ResNet-GAN, SNGAN and WGAN. Winning tickets consistently outperform random tickets on all models. Experiments are performed on $5$ random runs, and the error bars represent $+/-$ standard deviation across runs. }
\label{gan_plots}
\end{figure*}

\begin{figure*}[h]
\centering
\subfigure{\includegraphics[width=0.3\textwidth, trim={0.9cm 0.9cm 1cm 0.9cm}, clip]{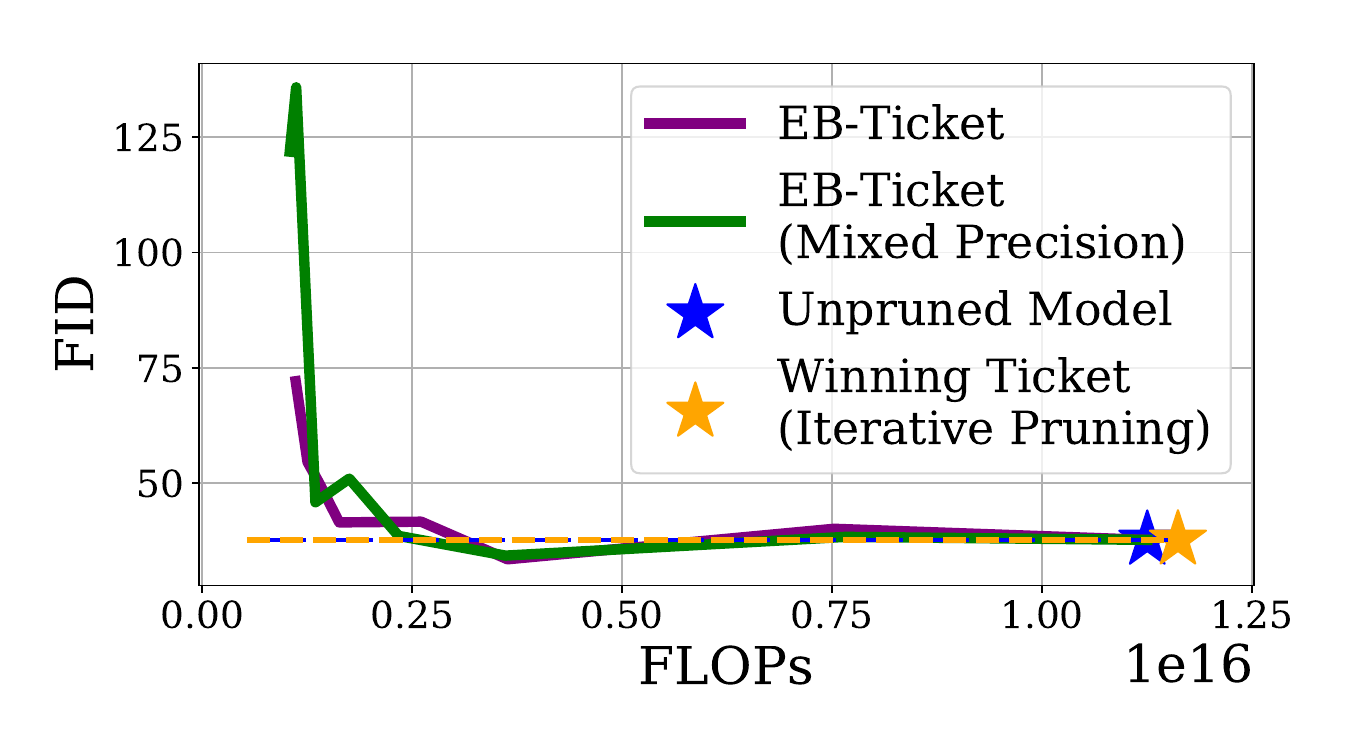}}
\subfigure{\includegraphics[width=0.3\textwidth, trim={1.3cm 0.9cm 1cm 0.9cm}, clip]{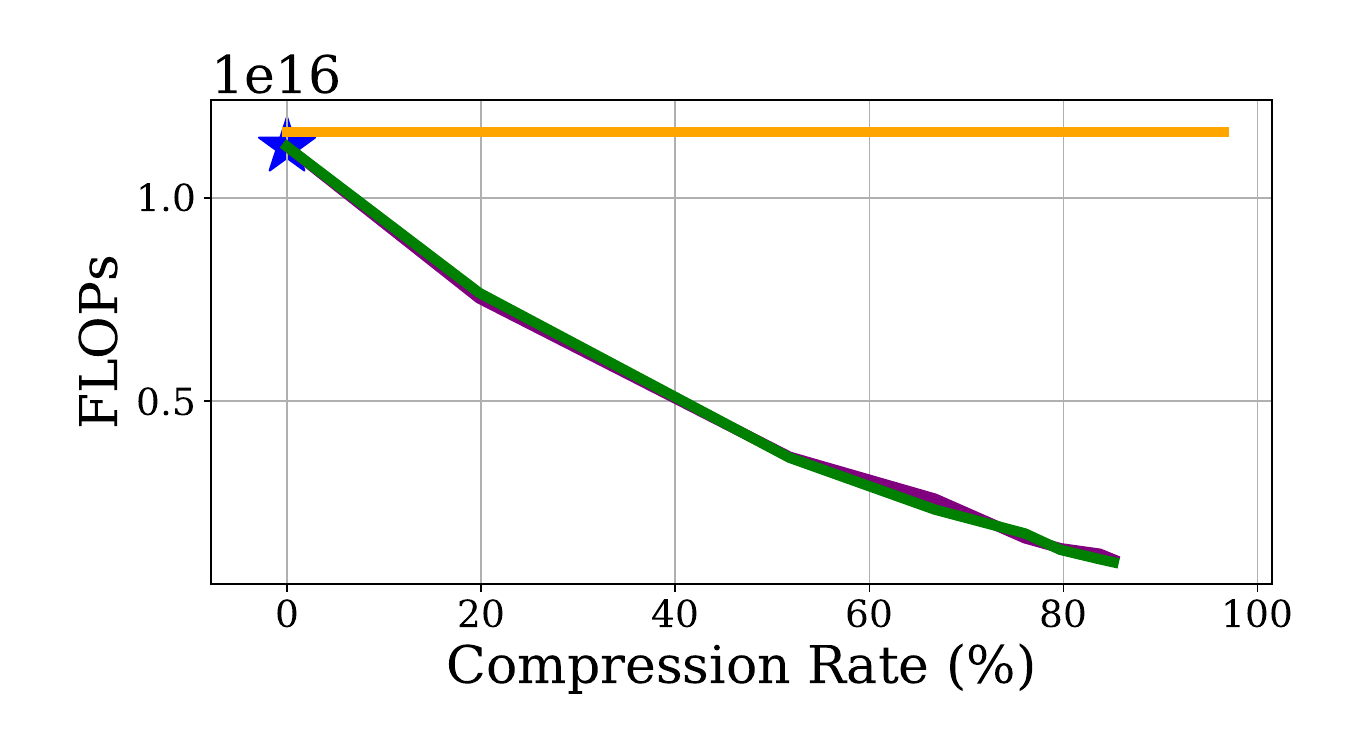}}
\subfigure{\includegraphics[width=0.3\textwidth, trim={0.9cm 0.9cm 1cm 0.9cm}, clip]{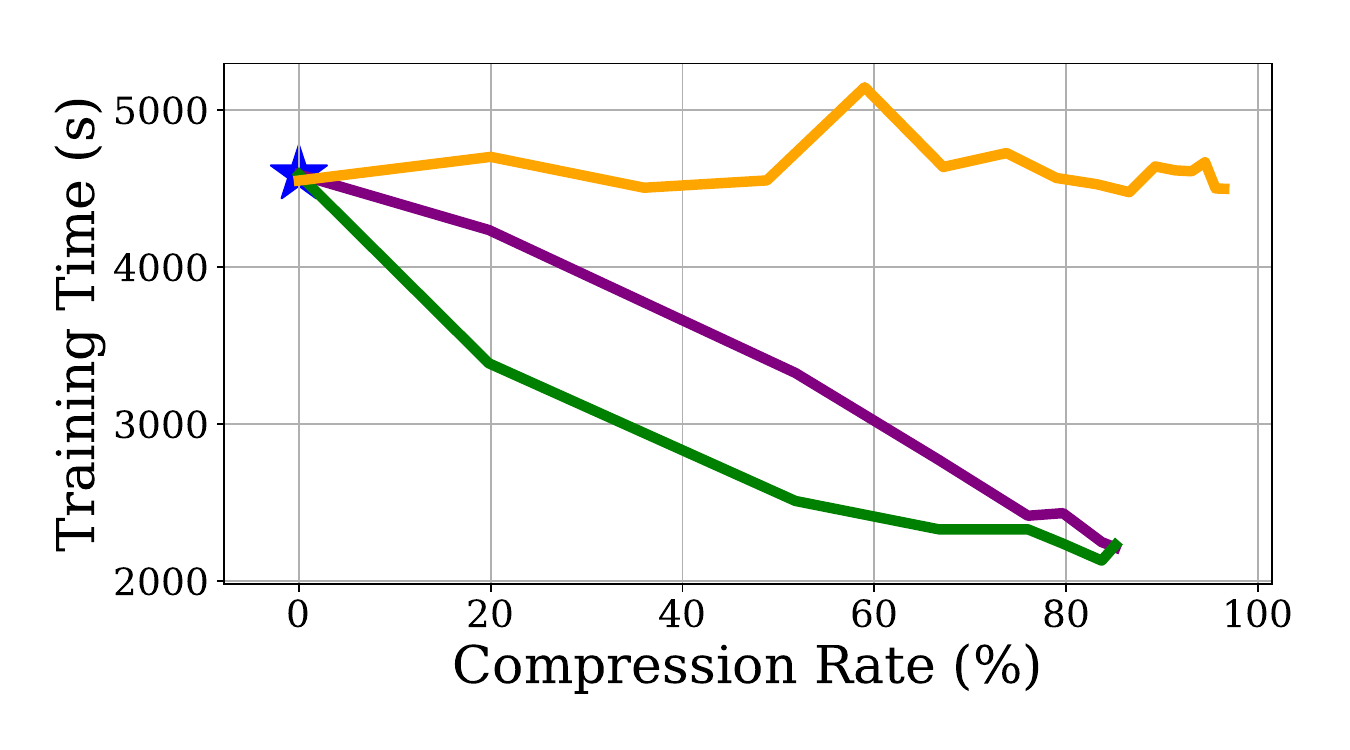}}
\caption{\textbf{Early-Bird Tickets} In the first plot, we compare the FID of tickets against FLOPs (floating-point operations). In the second plot, we see the change in FLOPs at different compression rates. EB-tickets show a significant reduction in FLOPs while maintaining performance. In the last plot, we see the change in training time of tickets at different compression rates. Mixed precision EB-tickets train the fastest. Experiments are performed on $5$ random runs, and the error bars represent $+/-$ standard deviation across runs.}
\label{eb_plots}
\end{figure*}

\section{Results}
\subsection{Winning Tickets in AutoEncoders}
In this section, we discuss the the winning tickets in AutoEncoders by evaluating the reconstruction loss with random tickets. In Figure \ref{auto_plots}, we observe that the winning tickets of a Linear AutoEncoder on MNIST preserve the reconstruction loss up to a sparsity of 89\%. The same tickets, when randomly initialized, perform visibly worse. In Convolutional AutoEncoders, we achieve winning ticket sparsity up to 96\% in CIFAR-10 and 99\% in Celeb-A. Note that a sparsity of 99\% reduces the number of weights from 3 million to around only 30K. 


It is also interesting to see how the winning ticket training curves are more well-behaved across runs compared to those of randomly-initialized tickets especially at higher pruning percentages. This indicates that good initializations do help stabilize the training process. Finally, the images reconstructed by the winning tickets maintain high quality, thus validating the presence of lottery tickets in AutoEncoders (See Appendix for more details). 

In Figure \ref{auto_plots}, we also demonstrate how differently the AutoEncoder behaves when we prune both components (Winning Ticket, green curve) and when a single component is pruned (blue, red curves). We observe that winning tickets with both components pruned, outperform single-component-pruned tickets. This implies that comparable network parameters in the encoder and decoder of AutoEncoders are essential for best results.

\subsection{Winning Tickets in Variational AutoEncoders}
The presence of winning tickets in AutoEncoders provides a strong indication that they could exist in VAEs as well. In VAEs, we observe that a sparsity of 79\% preserves FID in both CIFAR-10 and Celeb-A as shown in Figure \ref{vae_plots}. We also show that winning tickets are not restricted to the loss function or the architecture of the VAE. Winning tickets are visible in $\beta$-VAEs at 87\% sparsity and on ResNet-VAE at 93\% sparsity on both CIFAR-10 and Celeb-A. Winning tickets are also visible when evaluated on the classification accuracy and generated images (See Appendix for more details). It is evident that the ResNet-VAE shows winning tickets of highest quality in terms of FID, accuracy and generated images., effectively reducing the model size from 2.8 million to around only 196K. 

\begin{figure}[h]
\centering
\subfigure{\includegraphics[width=0.23\textwidth, trim={0.9cm 0.9cm 1cm 0.9cm}, clip]{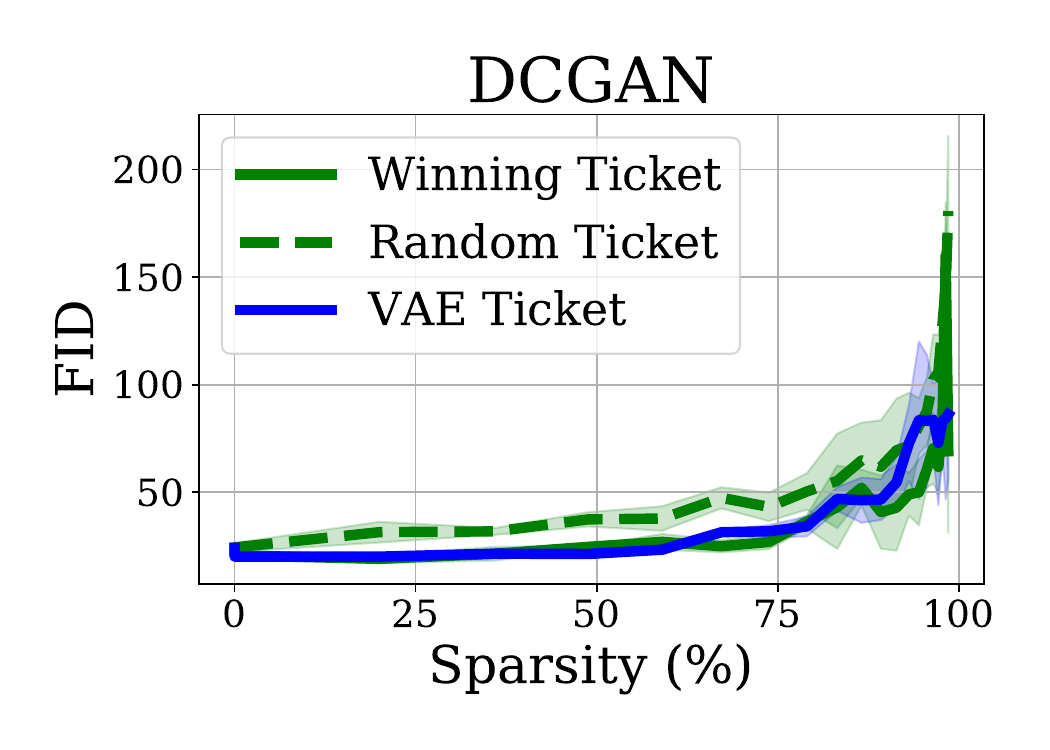}}
\subfigure{\includegraphics[width=0.23\textwidth, trim={0.9cm 0.9cm 1cm 0.9cm}, clip]{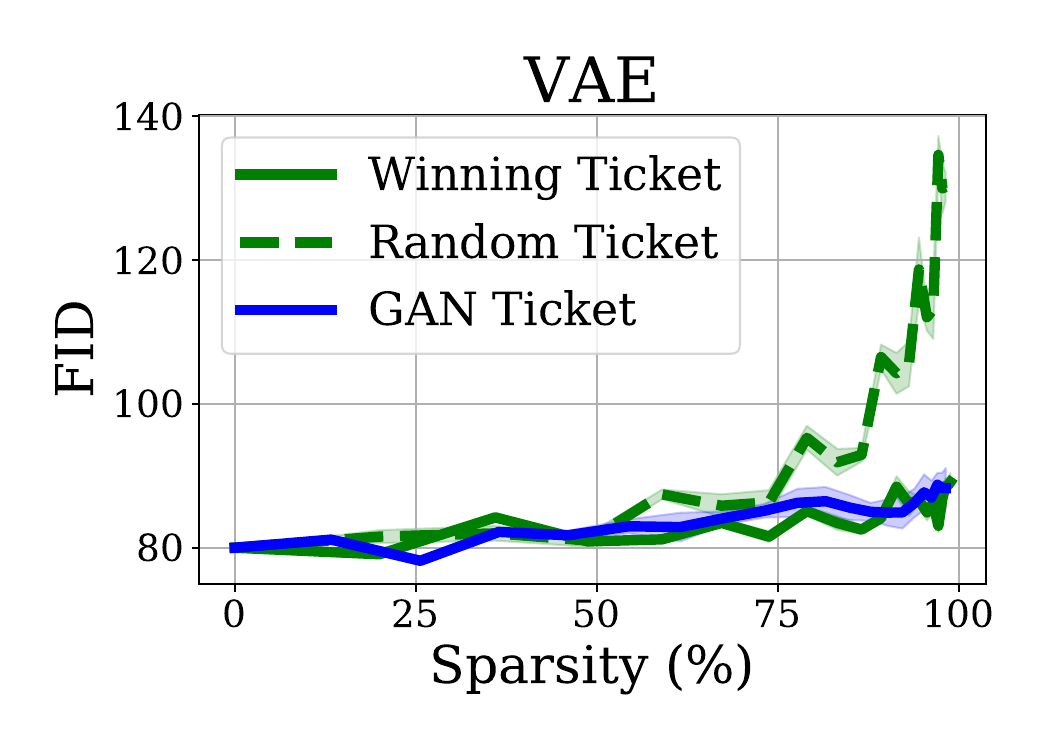}}
\caption{\textbf{Transferability of winning tickets.} The plot on the top compares DCGAN winning tickets against the VAE winning ticket (transfer ticket) trained on DCGAN. The performance of the transfer ticket is comparable to that of GAN's own winning ticket. The plot on the bottom shows the other direction where VAEs are trained with tickets obtained from the GAN. Performance of transfer ticket is on-par with VAE's own ticket. Experiments are performed on $5$ random runs, and the error bars represent $+/-$ standard deviation across runs.}
\label{transfer_plots}
\end{figure}

Figure \ref{vae_plots} also shows us the behavior of the VAE when single components are pruned. Note that we can only achieve around 50\% network sparsity when a single component of the network is pruned, since the other component accounts for the other 50\%. We observe that pruning the encoder shows negligible change in FID and is almost aligned with the winning ticket performance even up to 50\% network sparsity. Decoder pruning performs significantly worse than encoder pruning and the winning ticket. Our observation indicates that VAE can be trained with extremely sparse encoders, without affecting its performance. 

\subsection{Winning Tickets in Generative Adversarial Networks}
VAEs and AutoEncoders are both unsupervised models and their optimization objectives, based on minimizing the reconstruction and ELBO, is a minimization problem resembling that of the supervised prediction models. On the other hand, GANs are formulated as a min-max optimization problem which fundamentally differs from single minimization problems. In this section, we desire to see whether or not winning tickets exist in such generative models that are optimized game-theoretically.

In Figure \ref{gan_plots}, we see that tickets are seen at 83\% sparsity in DCGAN, 89\% in SNGAN and 79\% in ResNet-DCGAN. WGAN shows winning tickets of sparsity 73.7\% on CIFAR-10 and 59\% on Celeb-A (See Appendix for more details). The best performing winning tickets bring the size of the network from 6.7 million parameters to nearly 737K. We also see winning tickets when we evaluate the discriminator loss and the quality of generated images (See Appendix for more details). These results confirm the lottery ticket hypothesis in GANs under different loss functions, architectures and evaluation metrics. 

We also show in Figure \ref{gan_plots} that, similar to AutoEncoders, winning tickets when both components are pruned (green curve) outperform tickets with single component pruning (blue, red curves). With this observation, we conclude that the generator and discriminator in GANs should be of comparable sizes for GANs to perform well, even under very sparse regimes. 

\subsection{Transferability of Winning Tickets from VAEs to GANs}
In this section, we show evidence of transferability of winning tickets across generative models. We know that the VAE's decoder network and GAN's generator network share the same architecture and task. Therefore, we transfer VAE's decoder winning tickets and train them under a DCGAN generator setup while keeping the discriminator unpruned. The first plot in Figure \ref{transfer_plots} compares three cases: (1) The green and  (2) dashed green curves represent winning and random tickets found by pruning the DCGAN generator. (2) The blue curve shows the behavior of winning tickets when transferred from the VAE's decoder to the DCGAN's generator. We see that the VAE-initialized winning tickets match the performance of the tickets found using GANs, and preserve performance up to 80\%. We observe similar results when we transfer winning tickets from GAN's generator to the VAE decoder where the transferred tickets behave comparably to the VAE winning tickets. This shows that a single initialization succeeds in training winning tickets in both networks. Hence, this confirms our hypothesis that winning tickets can be transferred across different deep generative models and provides evidence for a universal weight initialization that could work well across a range of generative models.

\subsection{Early-Bird Tickets}
In this section, we observe the behavior of early-bird tickets and compare it with winning lottery tickets found using IMP. In Figure \ref{eb_plots}, (1) the purple line represents EB-tickets at different network compression rates ranging from 20\% to 95\%, (2) the green line represents the same EB-tickets under mixed-precision training, (3) the yellow star/line represents winning lottery tickets found in the previous sections and (4) the blue star represents the unpruned network. We observe that, in DCGAN, the training FLOPs can be reduced from 11.2 quadrillion (unpruned network) to 1.3 quadrillion (over 88\% reduction), with a negligible change in FID. Mixed-precision trained EB-tickets perform on par with full-precision trained EB-tickets in terms of FLOPs. The training time of mixed-precision EB-tickets (53.5\% reduction), however, is better than full-precision tickets (47\% reduction). This indicates that reduced-precision training is a simple strategy that can reduce memory and computation without compromising on the model performance. 

\begin{table}[h]
\caption{Comparing the performance of Early-Bird Tickets in DCGAN to other early pruning techniques}
\centering
\resizebox{0.48\textwidth}{!}{
 \begin{tabular}{l|c|c|c|c} 
 \toprule
 \textbf{\multirow{2}{*}{Pruning Technique}} & \textbf{\multirow{2}{*}{FID}} & \textbf{Number of} & \textbf{\multirow{2}{*}{FLOPs}} & \textbf{Training Time} \\[0.5ex]
  & & \textbf{weights} & & \textbf{(seconds)} \\
 \midrule
 \midrule
 Unpruned Network & 37.71 & 5651584 & 1.12e+16 & 4590.5 \\
 Iterative Magnitude Pruning & 37.77 & 5651584 & 1.16e+16 & 4477.7 \\
 EB-Ticket & \textbf{33.49} & \textbf{1148190} & \textbf{0.13e+16} & \textbf{2417.07} \\
 EB-Ticket (mixed-precision) & 34.28 & 1148190 & 0.13e+16 & 2131.09 \\
 SNIP & \multirow{2}{*}{56.02} & \multirow{2}{*}{5651584}  & \multirow{2}{*}{1.12e+16} & \multirow{2}{*}{4689.5} \\
 \cite{lee2018snip} & & & & \\
 GraSP & \multirow{2}{*}{65.53} & \multirow{2}{*}{5651584} & \multirow{2}{*}{1.12e+16} & \multirow{2}{*}{4603.3} \\
 \cite{Wang2020Picking} & & & & \\
 \bottomrule
\end{tabular} 
}
\label{comp_table}
\end{table}

The winning lottery tickets, on the other hand, show an increase in FLOPs compared to the unpruned network due to repetitive masking of the network after every iteration. The FLOPs and training time for winning lottery tickets are also consistently high across all sparsities, while EB-tickets consistently reduce FLOPs and training time as the compression rate increases. 

Finally, we compare the performance of EB-tickets to other recent early pruning strategies like SNIP \cite{lee2018snip} and GraSP \cite{Wang2020Picking} in Table \ref{comp_table}. SNIP and GraSP are sparse pruning strategies that prune the network at initialization. Therefore, although they prune networks very early in training, they show no reduction in FLOPs, training time and number of weights. More importantly, they are unfavorable for generative models as they do not produce good FIDs. The EB-tickets therefore, outperform SNIP and GraSP in every aspect. These results also align with recent work \cite{frankle2020pruning} that shows pruning at initialization is always inadequate.  

\section{Conclusion}
The key finding in this paper is that the Lottery Ticket hypothesis holds in deep generative models such as VAEs and GANs under different loss functions and architectures. We show that these winning tickets are visible under multiple evaluation metrics. We confirm that winning tickets can be transferred across generative models with different objective functions indicating that a single initialization can successfully train multiple generative models to convergence. Finally, with early-bird tickets we show the most effective and practical approach to train generative models using significantly lesser resources. Thus, large generative models can be optimized using lottery tickets with improved training time, storage and computation resources. Applying our findings on even larger GANs like BigGAN \cite{brock2018large}, is a direction for future research. 

\section{Acknowledgements}
This project was supported in part by NSF CAREER AWARD 1942230 and a Simons Fellowship on Deep Learning Foundations.

\section*{Broader Impact}
Deep generative models are used for a variety of tasks such as image generation, image editing, 3D object generation, video prediction and image in-painting. Powerful GANs that perform such tasks on large-scale datasets such as ImageNet \cite{imagenet_cvpr09}, require TPUs of 128 to 512 cores to generative high quality images. However, in reality, users capture, store and interact with images on mobile phones which have a limited compute power. Several learning tasks as mentioned above, therefore, become too far-fetched to accomplish in real-time on end-user devices. Our work opens up a possibility of training deep generative models without the requirement of powerful GPUs and large amounts of memory, thus potentially opening their applications to a broader community of users. We thus shift the focus to powerful initializations of small networks to achieve improved results. Finally, to the best of our knowledge, this work will not create any negative societal or ethical impacts.

\bibliography{main}

\clearpage
\newpage
\appendix
\renewcommand\thefigure{\thesection.\arabic{figure}}
\section{Appendix}
\setcounter{figure}{0}    
\subsection{Winning Tickets in Generative Adversarial Networks on Celeb-A}
\begin{figure}[h]
\centering
\subfigure{\includegraphics[width=0.23\textwidth, trim={0.9cm 0.9cm 1cm 0.9cm}, clip]{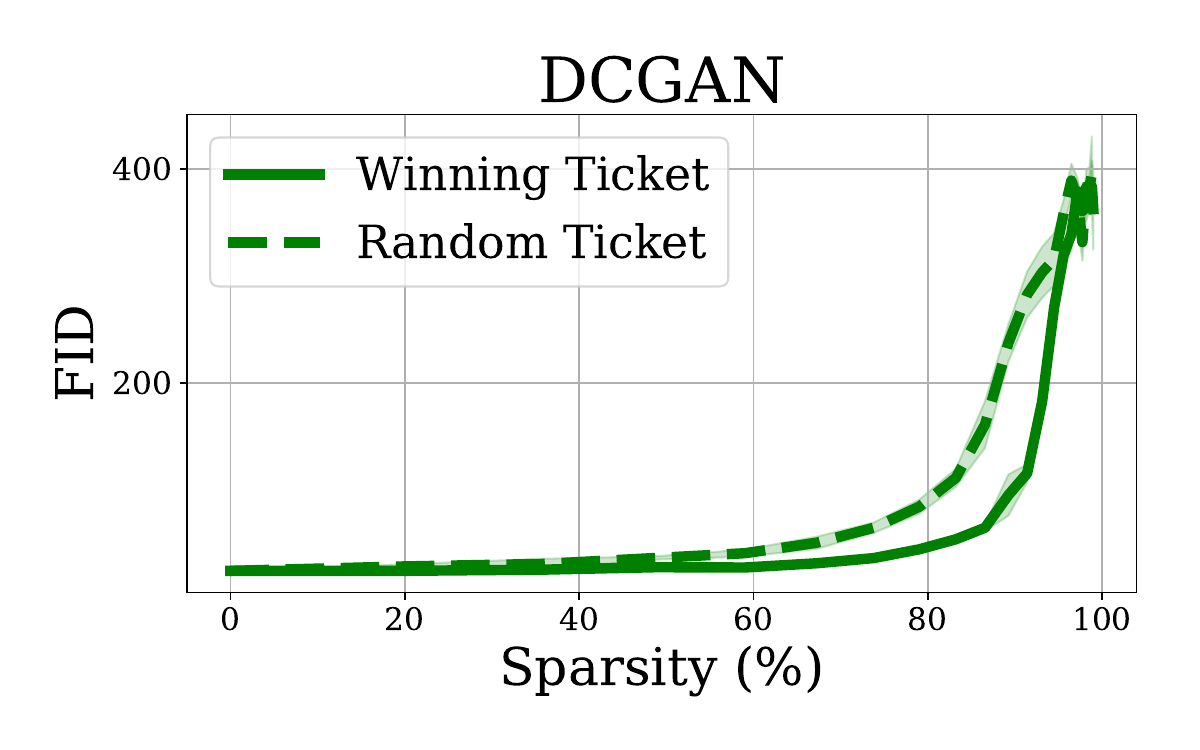}}
\subfigure{\includegraphics[width=0.23\textwidth, trim={0.9cm 0.9cm 1cm 0.9cm}, clip]{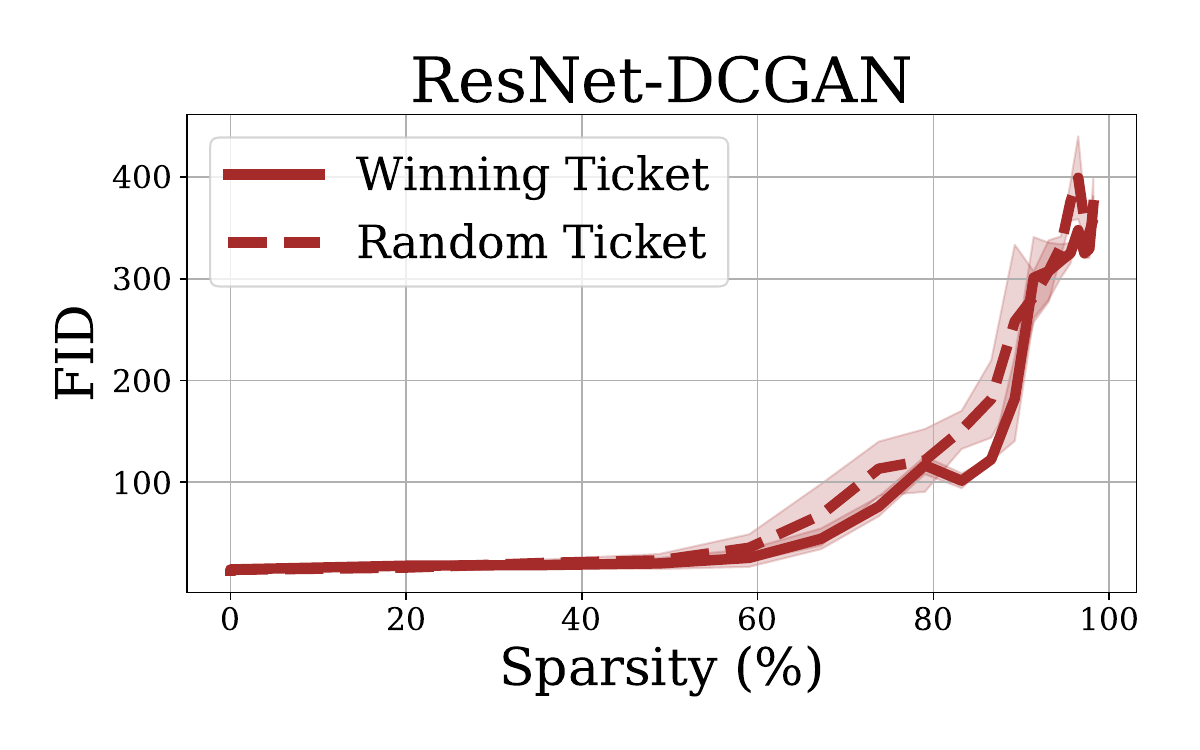}}
\subfigure{\includegraphics[width=0.23\textwidth, trim={0.9cm 0.9cm 1cm 0.9cm}, clip]{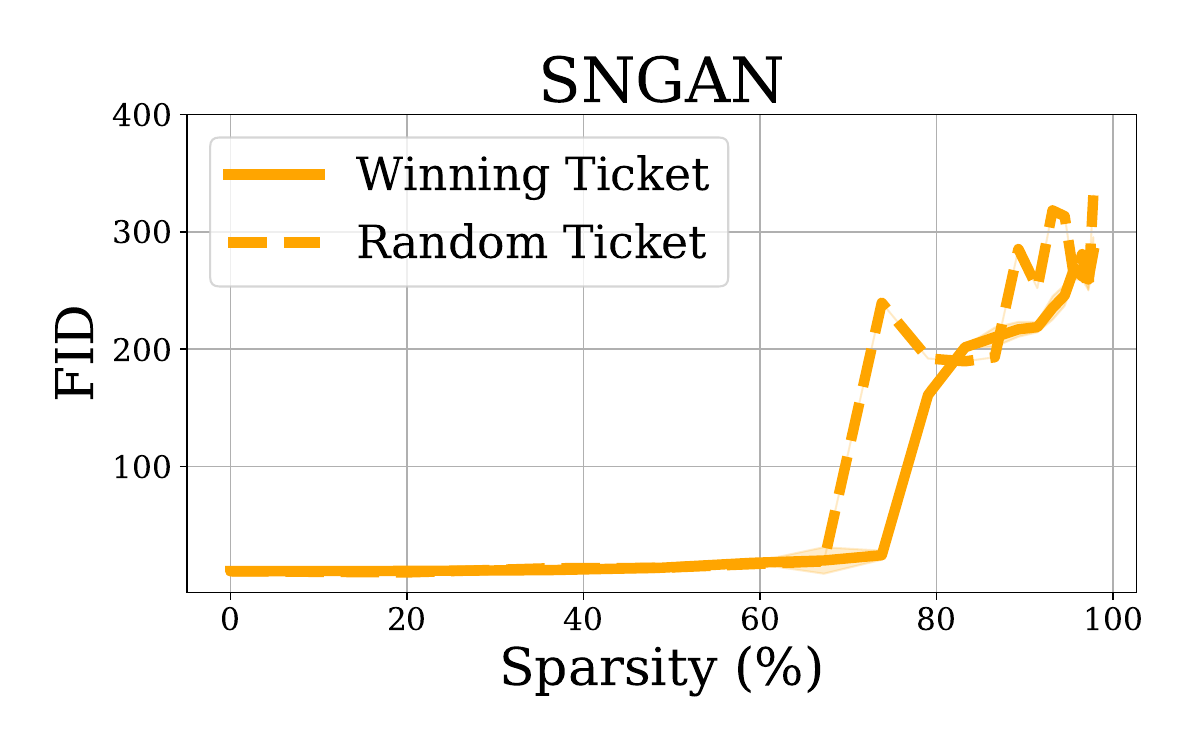}}
\subfigure{\includegraphics[width=0.23\textwidth, trim={0.9cm 0.9cm 1cm 0.9cm}, clip]{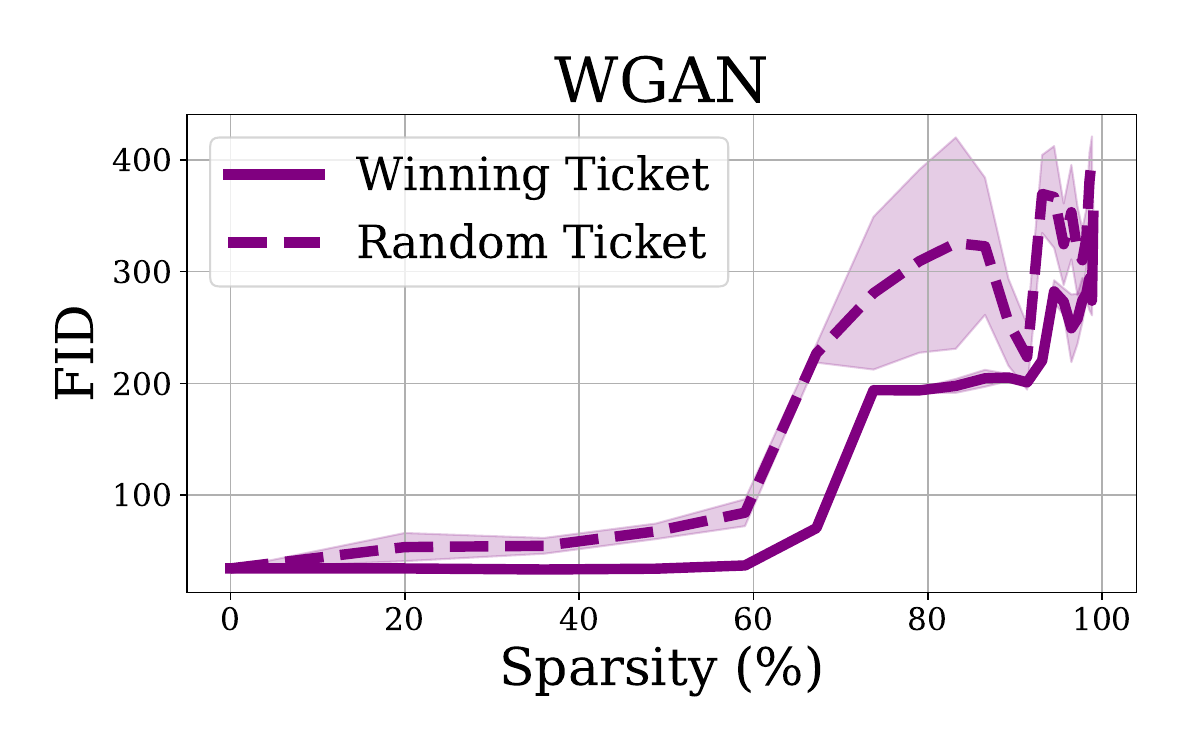}}
\caption{\textbf{GAN experiments. } We plot the FID scores of winning tickets and random tickets of $4$ GAN models trained on Celeb-A dataset: DCGAN, ResNet-GAN, SNGAN and WGAN. Winning tickets consistently outperform random tickets on all settings. Experiments are performed on $5$ random runs, and the error bars represent $+/-$ standard deviation across runs. }
\label{gan_celeba_plots}
\end{figure}

In Figure \ref{gan_celeba_plots} we plot the FID of winning and random tickets trained on $4$ GAN models: DCGAN, Spectral Normal GAN, Wasserstein GAN and ResNet-DCGAN on the Celeb-A dataset. We consistently observe that winning tickets preserve FID to a larger extent than random tickets. We show that both DCGAN and SNGAN show winning tickets at 80\% sparsity, WGAN at 50\% and ResNet-GAN at 75\% sparsity. Therefore, the lottery ticket hypothesis holds on GANs trained on Celeb-A, across loss functions and architectures. 

\begin{figure}[h]
\centering
\subfigure{\includegraphics[width=0.2\textwidth, trim={0.9cm 0.9cm 1cm 0.9cm}, clip]{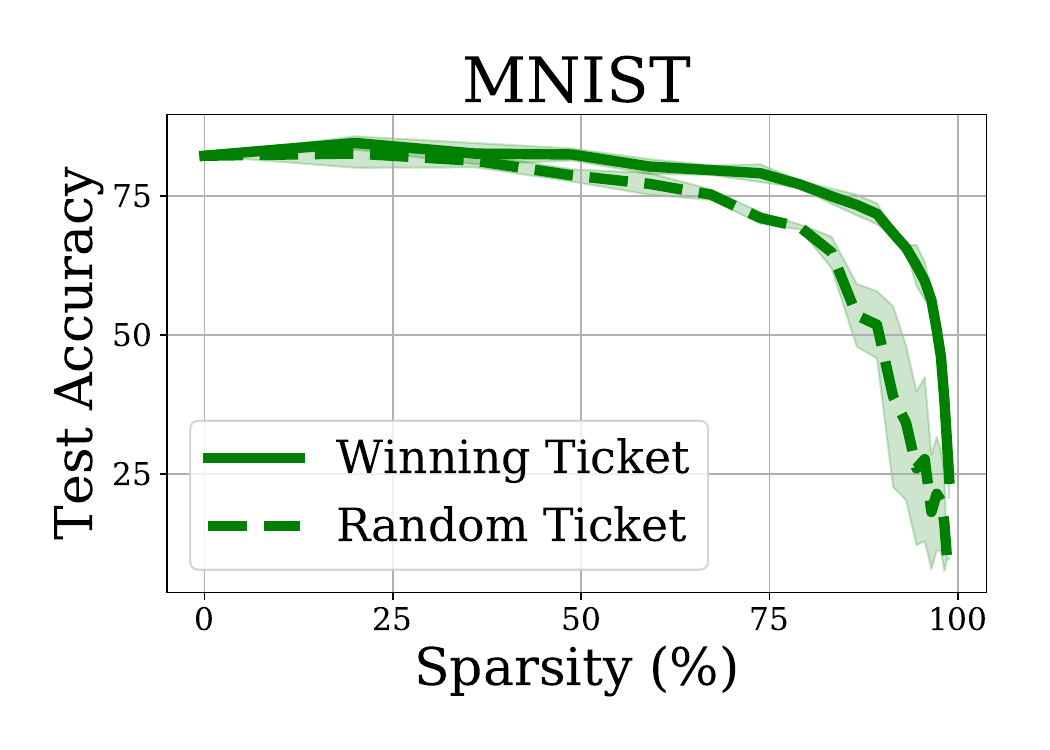}}
\subfigure{\includegraphics[width=0.2\textwidth, trim={0.9cm 0.9cm 1cm 0.9cm}, clip]{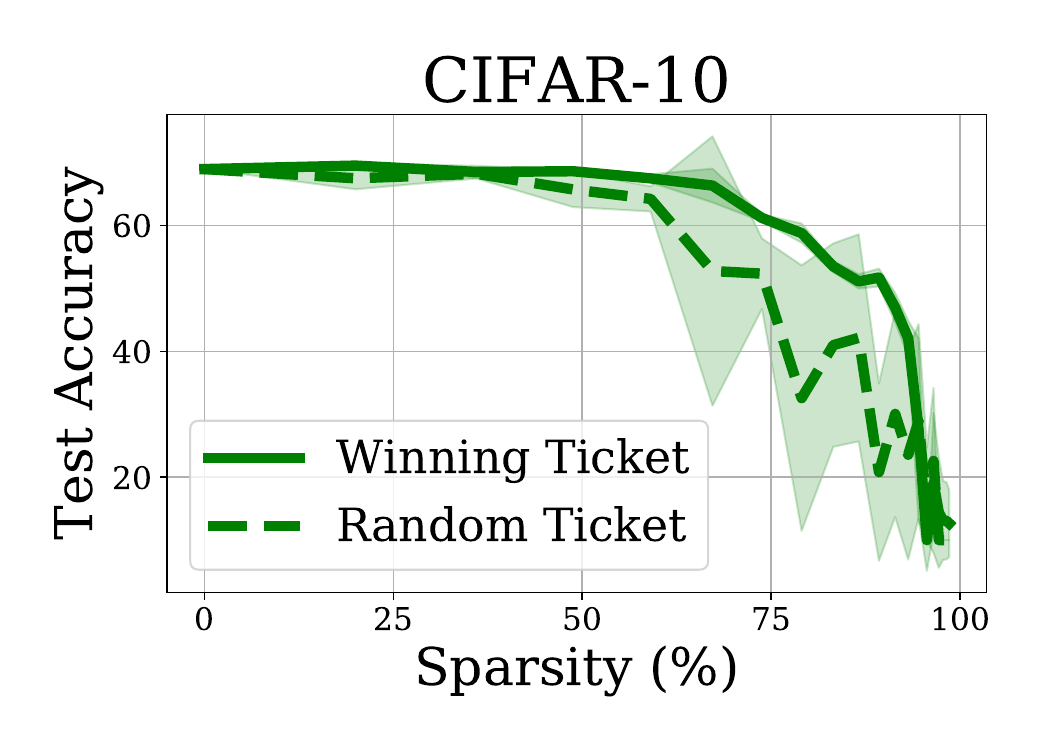}}
\subfigure{\includegraphics[width=0.2\textwidth, trim={0.9cm 0.9cm 1cm 0.9cm}, clip]{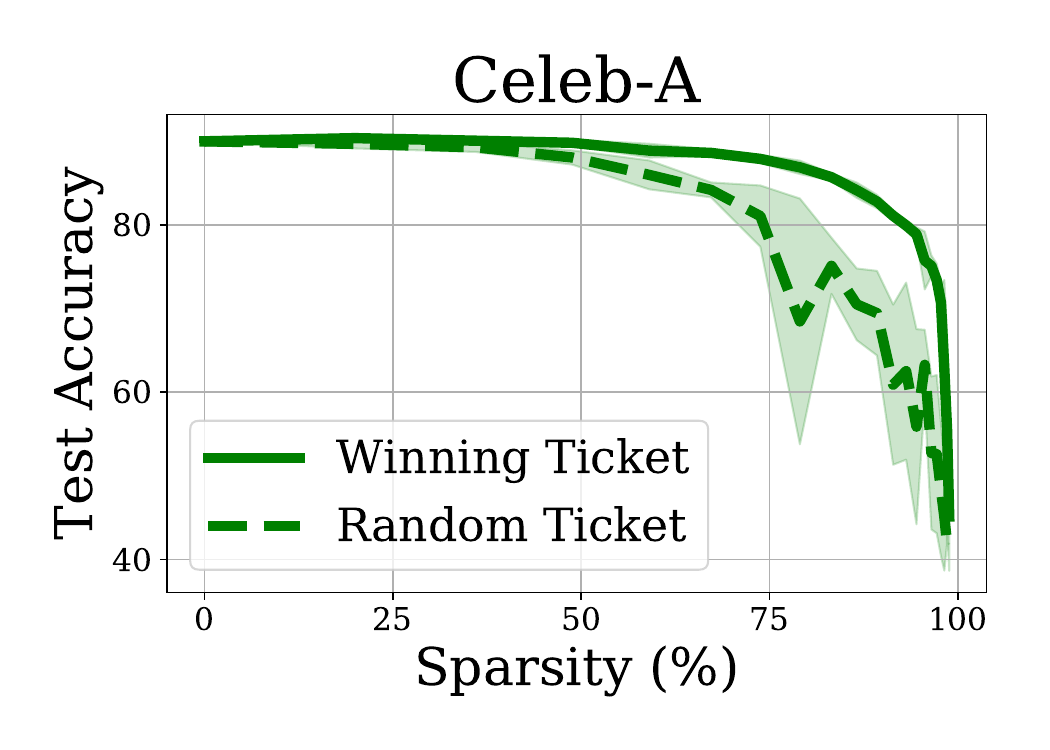}}
\caption{\textbf{AutoEncoder experiments.} We plot downstream classification accuracy of winning and random tickets. Performance of winning tickets are consistently better than random tickets. Each experiment is performed on $5$ random runs, and the error bars represent $+/-$ standard deviation across runs.}
\label{auto_acc_plots}
\end{figure}

\subsection{Evaluating Winning Tickets using Classification Accuracy in AutoEncoders and VAEs}
\begin{figure}[h]
\centering
\subfigure{\includegraphics[width=6.9cm, trim={0.9cm 0.9cm 1cm 0.9cm}, clip]{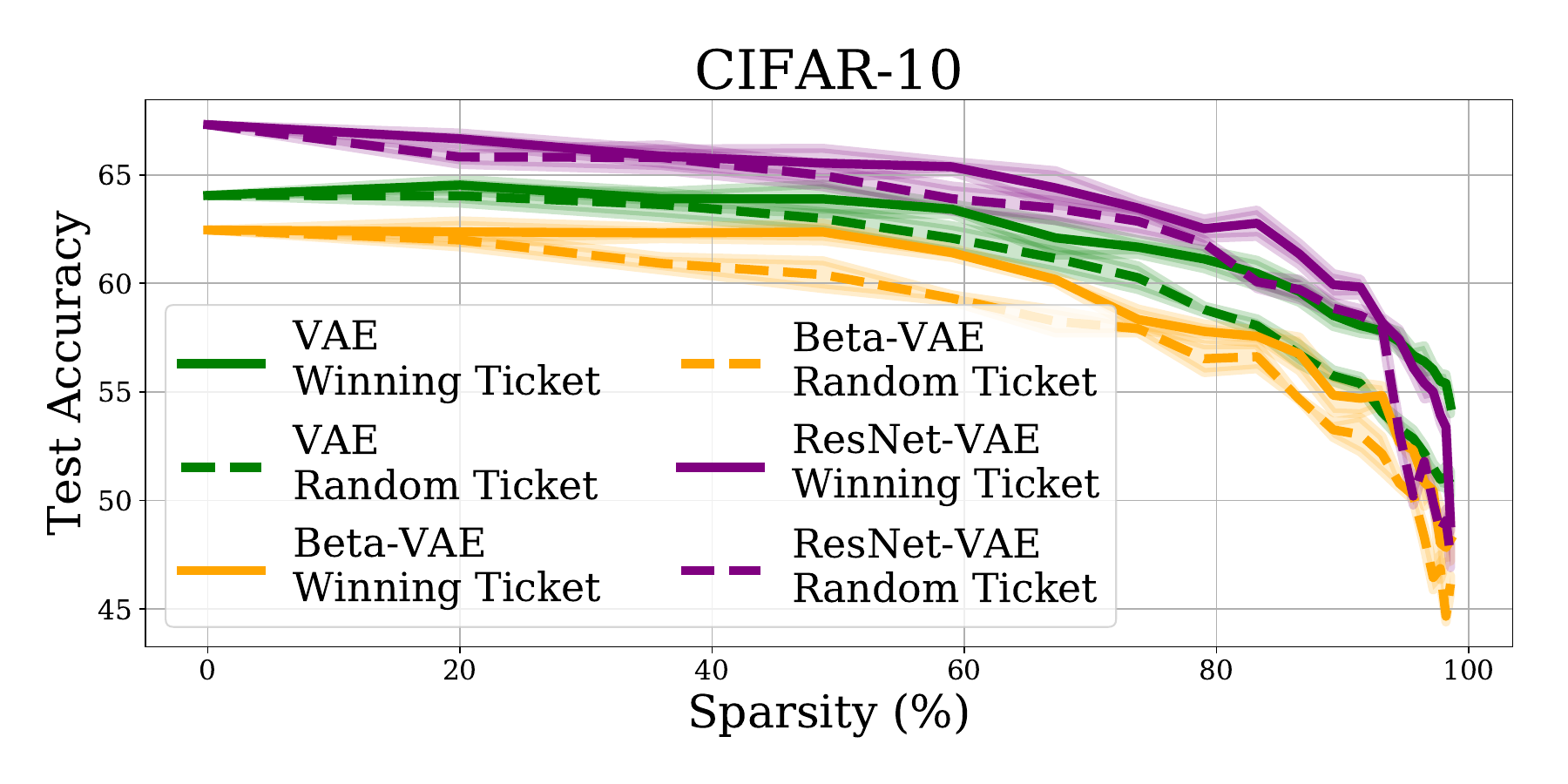}}
\subfigure{\includegraphics[width=6.9cm, trim={0.9cm 0.9cm 1cm 0.9cm}, clip]{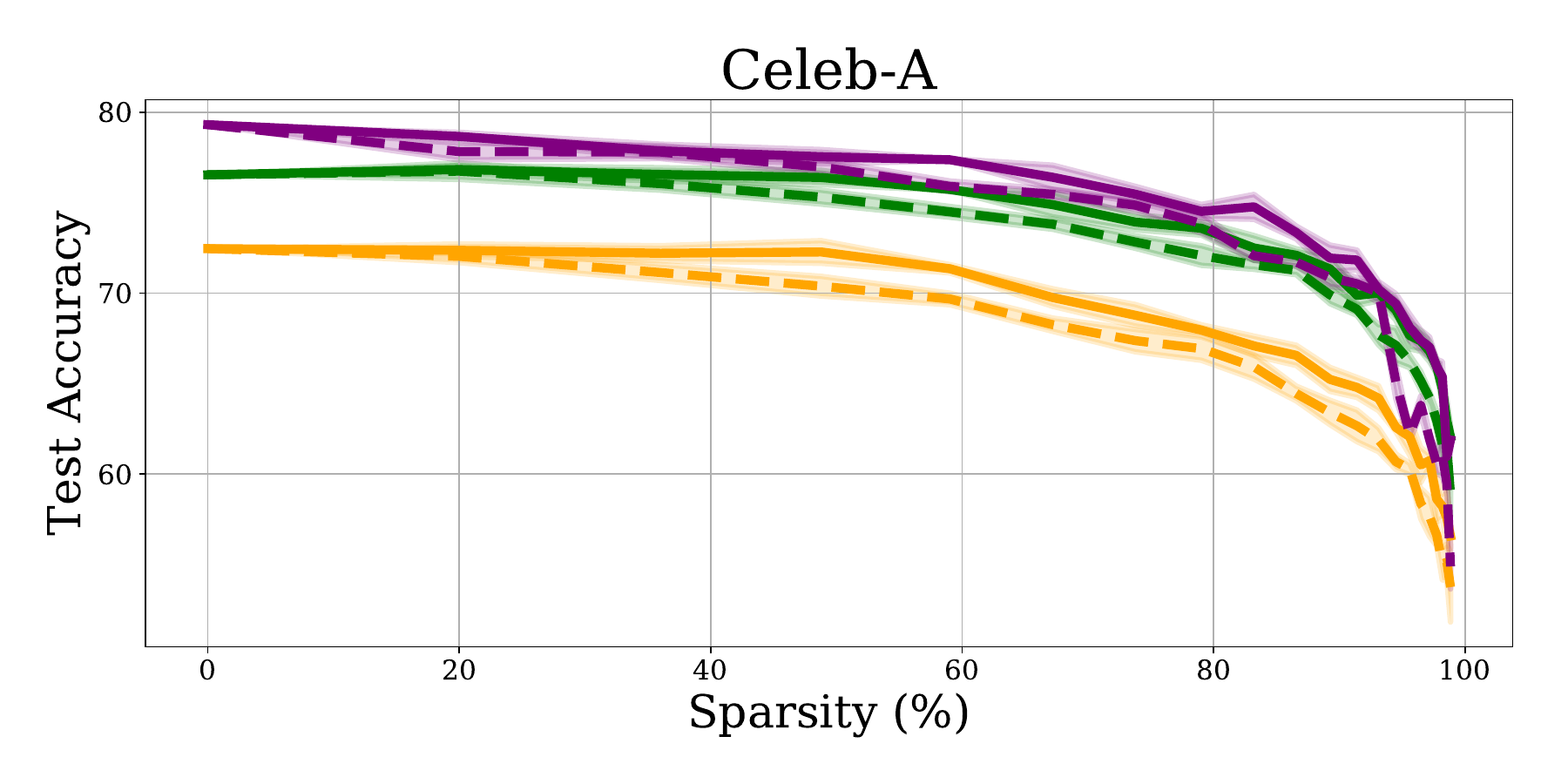}}
\caption{\textbf{VAE experiments.} We show downstream classification accuracy on CIFAR-10 and Celeb-A. Results are reported on three models: VAE, $\beta$-VAE, and ResNet-VAE. Winning tickets outperform random tickets on all models. Experiments are performed on $5$ random runs, and the error bars represent $+/-$ standard deviation across runs.}

\label{vae_acc_plots}
\end{figure}

In Figure \ref{auto_acc_plots}, we show the downstream classification test accuracy of the AutoEncoder on MNIST, CIFAR-10 and Celeb-A. We observe that winning tickets consistently out-perform random tickets with better accuracy for all sparsity levels. Similarly, in Figure \ref{vae_acc_plots}, we observe the same behavior where winning tickets show better classification accuracy of generated images in VAE, $\beta$-VAE and ResNet-VAE. We thus show that winning tickets are visible in AutoEncoders and VAEs using multiple evaluation metrics. 

\begin{figure}[h]
\centering
\subfigure{\includegraphics[width=0.23\textwidth]{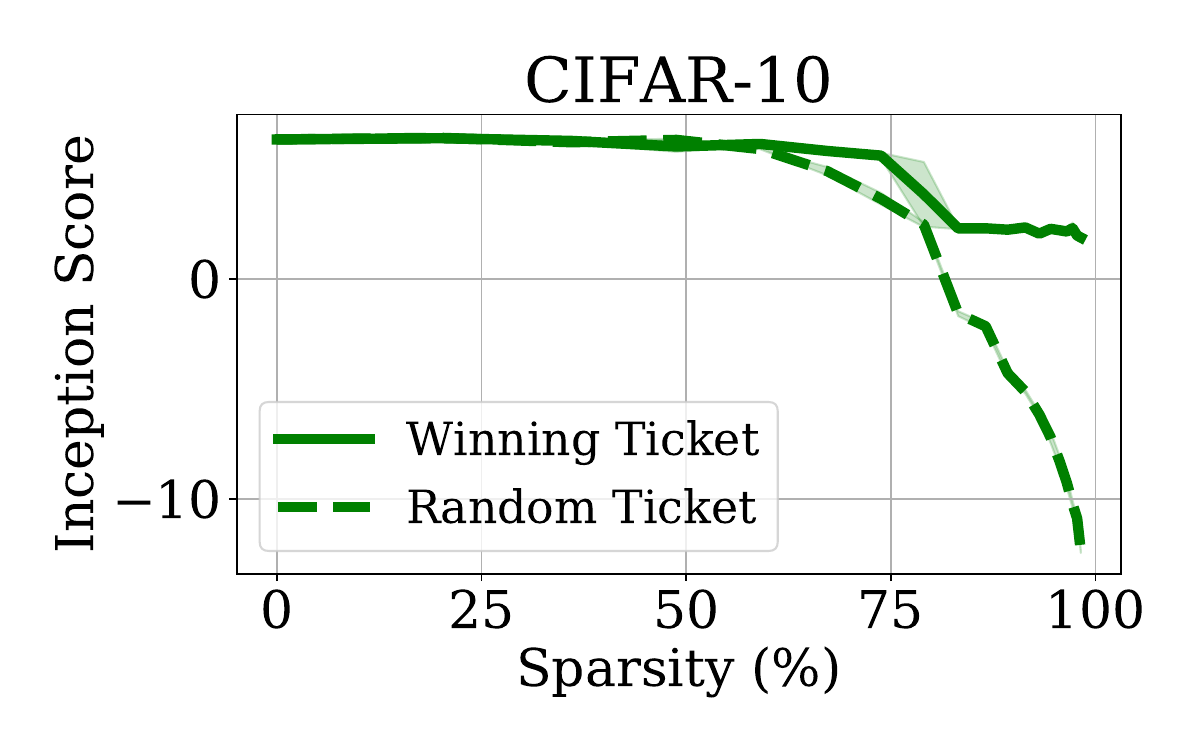}}
\subfigure{\includegraphics[width=0.23\textwidth]{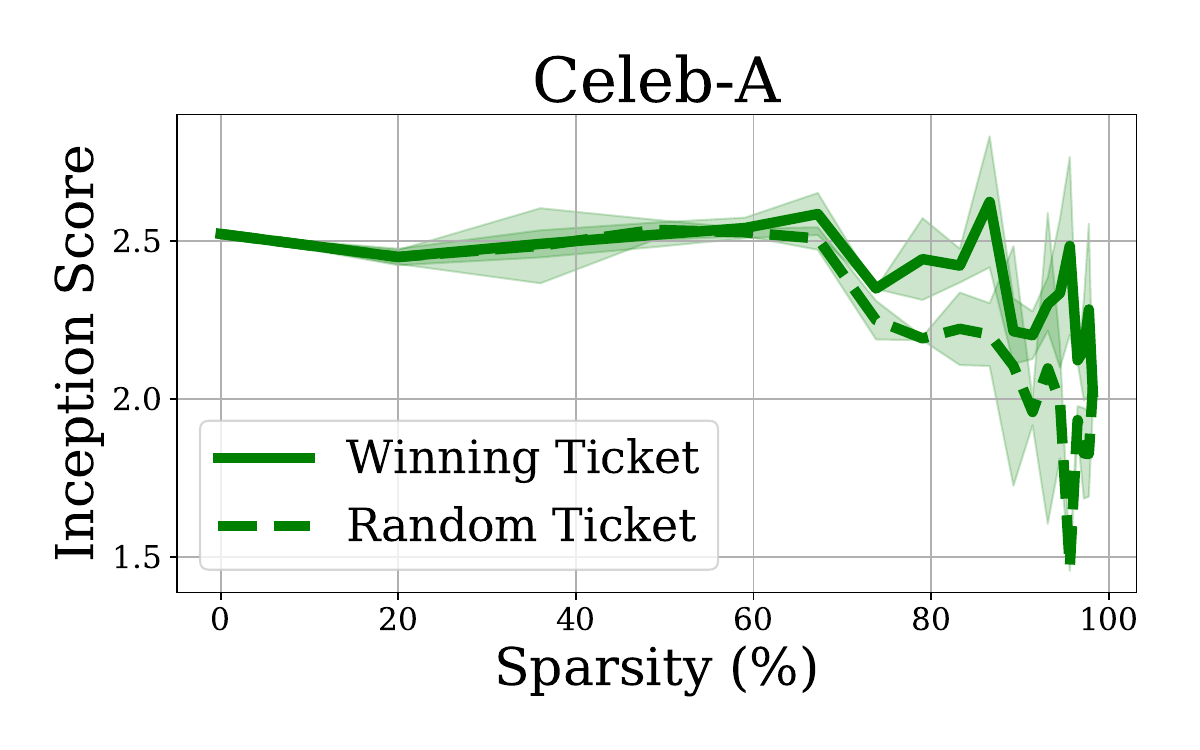}}
\caption{\textbf{Inception Score. } We plot the inception score of winning and random tickets of DCGAN trained on CIFAR-10 and Celeb-A. Winning tickets consistently show higher inception scores. Experiments are performed on $5$ random runs, and the error bars represent $+/-$ standard deviation across runs.}
\label{gan_inception_plots}
\end{figure} 

\subsection{Evaluating Winning Tickets using Discriminator Loss and Inception Score in GANs}
\begin{figure}[h]
\centering
\subfigure{\includegraphics[width=0.23\textwidth]{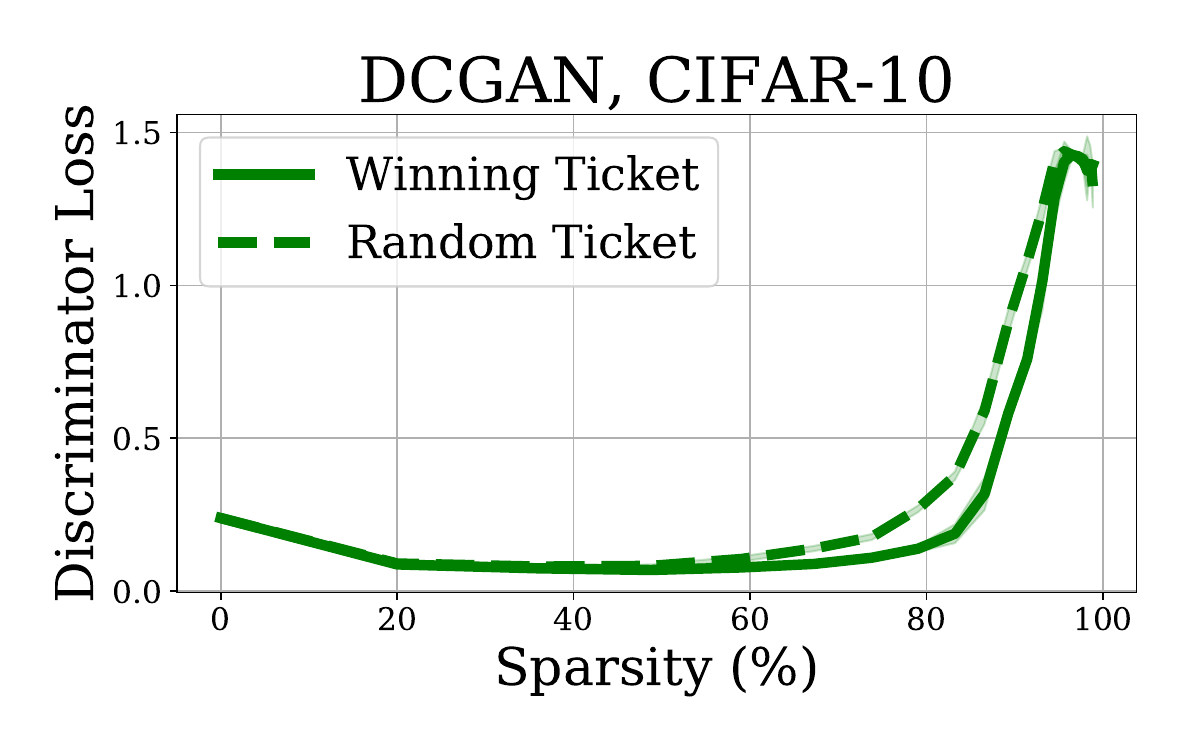}}
\subfigure{\includegraphics[width=0.23\textwidth]{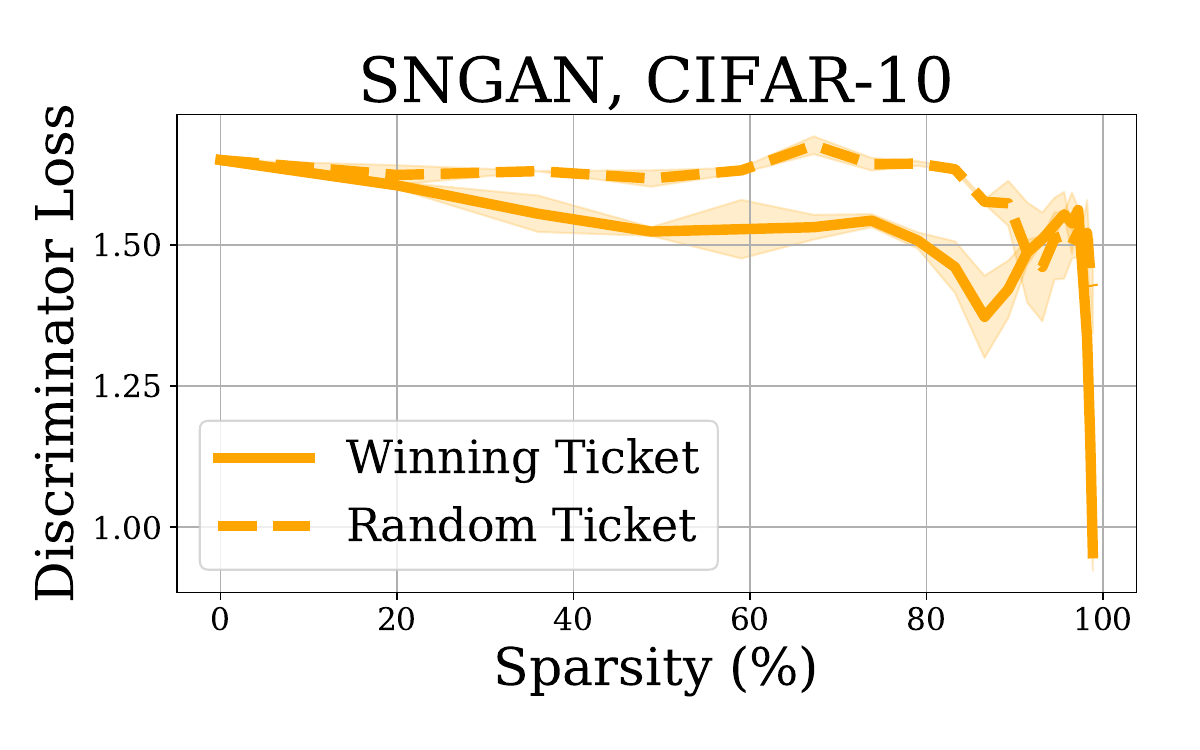}}
\subfigure{\includegraphics[width=0.23\textwidth]{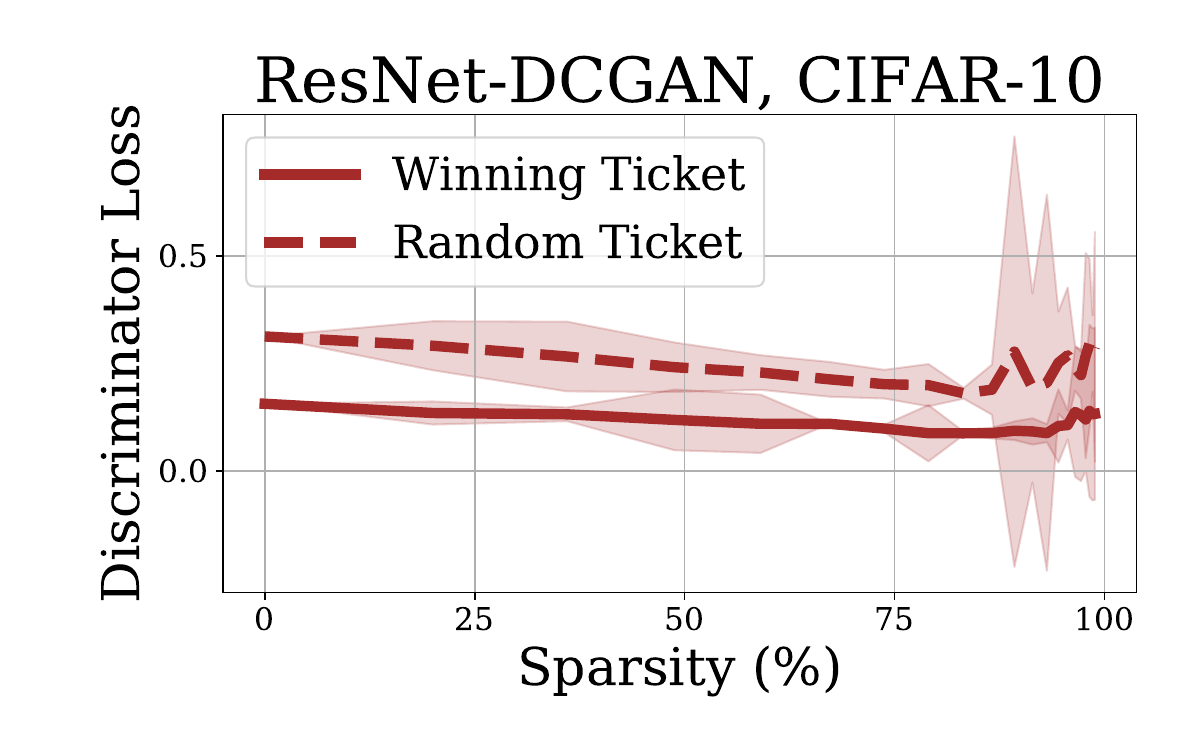}}
\subfigure{\includegraphics[width=0.23\textwidth]{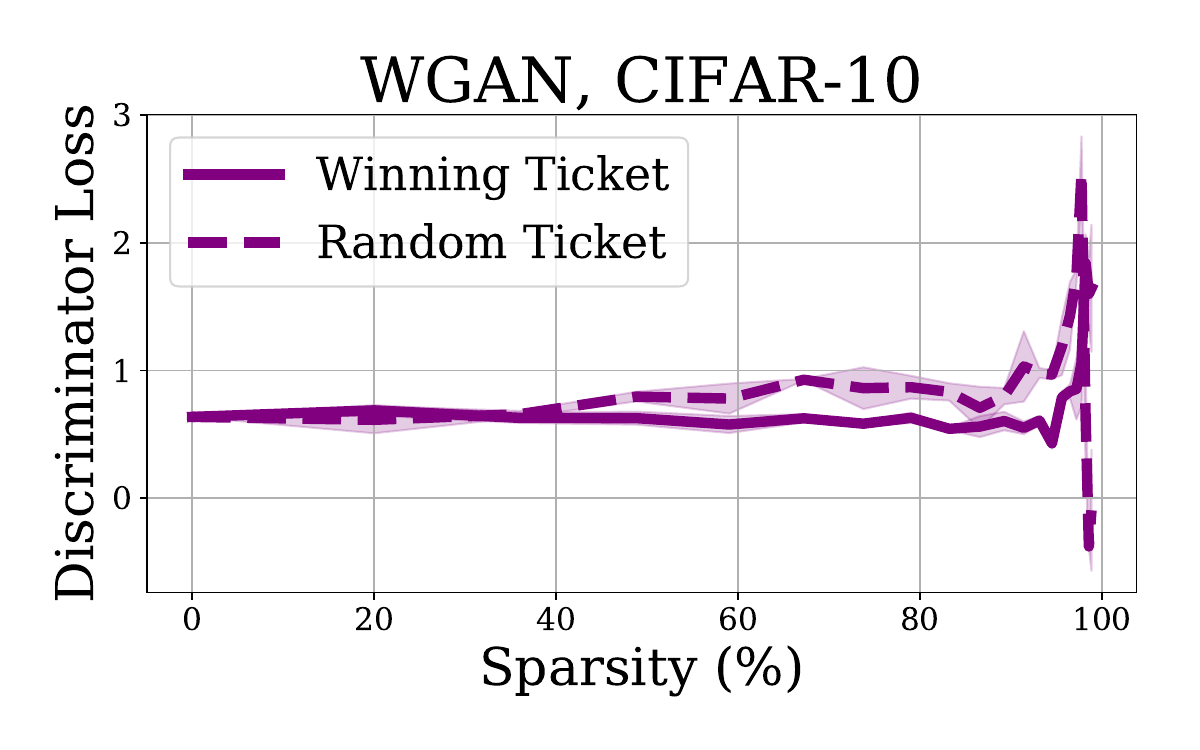}}
\subfigure{\includegraphics[width=0.23\textwidth]{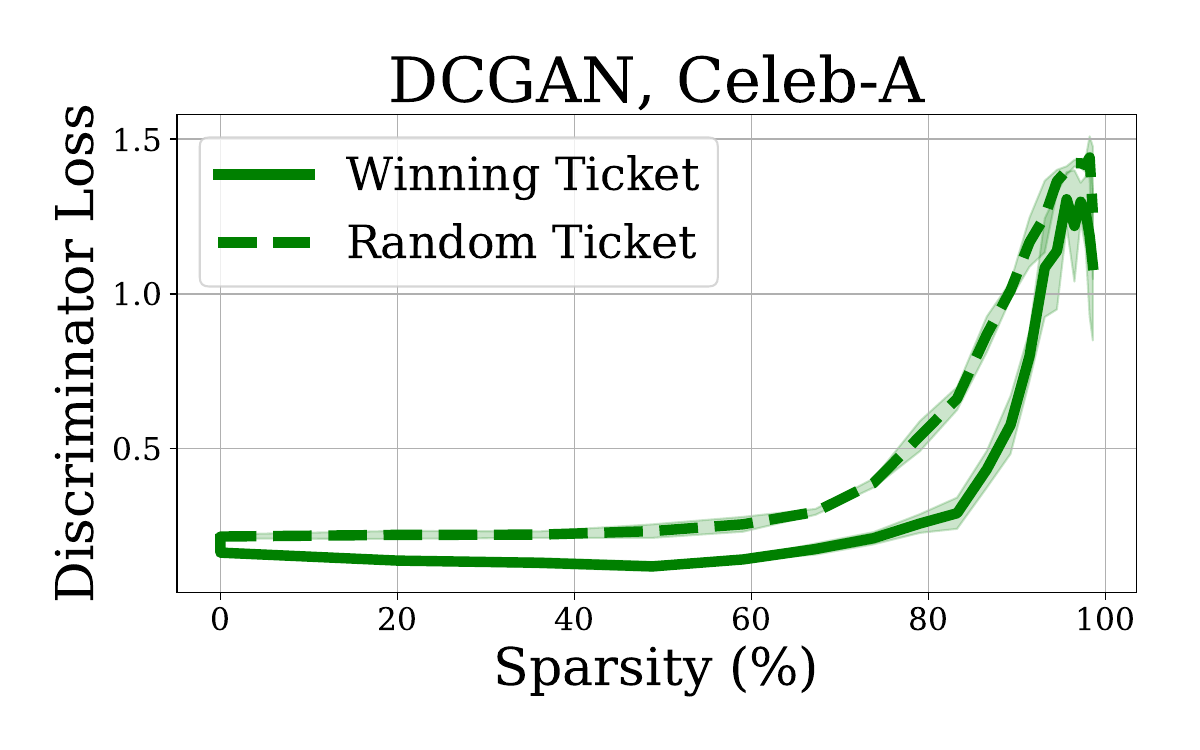}}
\subfigure{\includegraphics[width=0.23\textwidth]{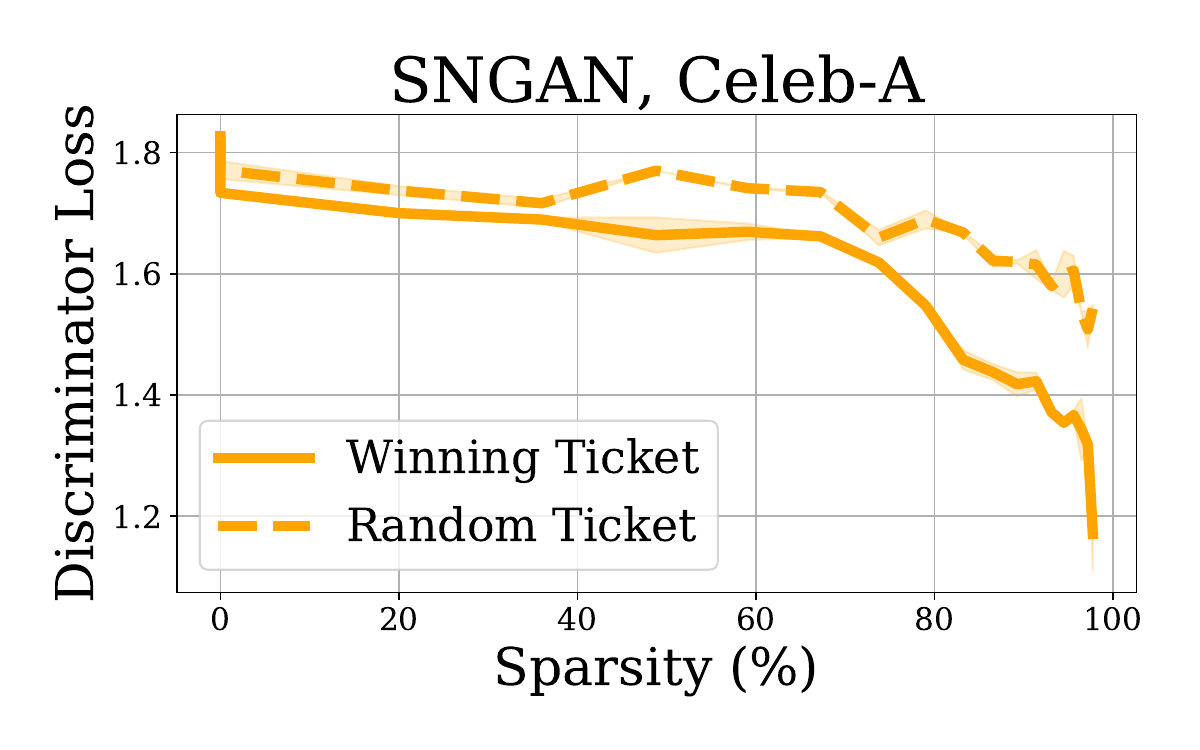}}
\subfigure{\includegraphics[width=0.23\textwidth]{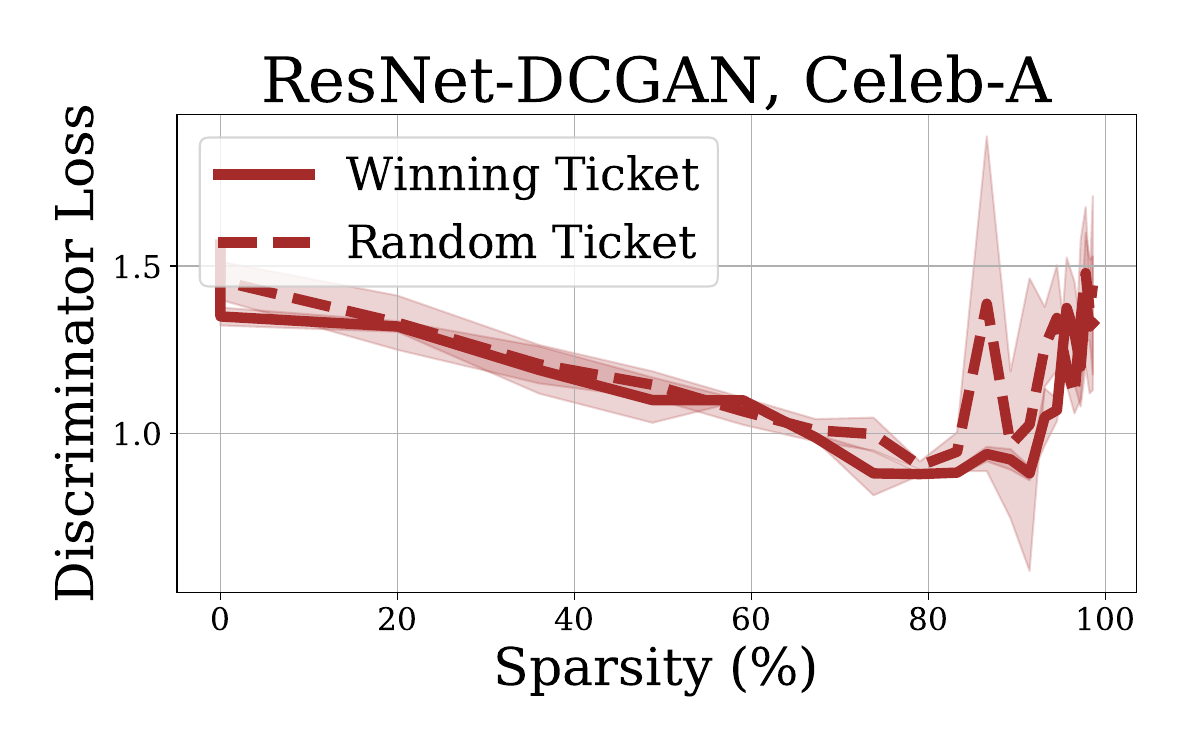}}
\subfigure{\includegraphics[width=0.23\textwidth]{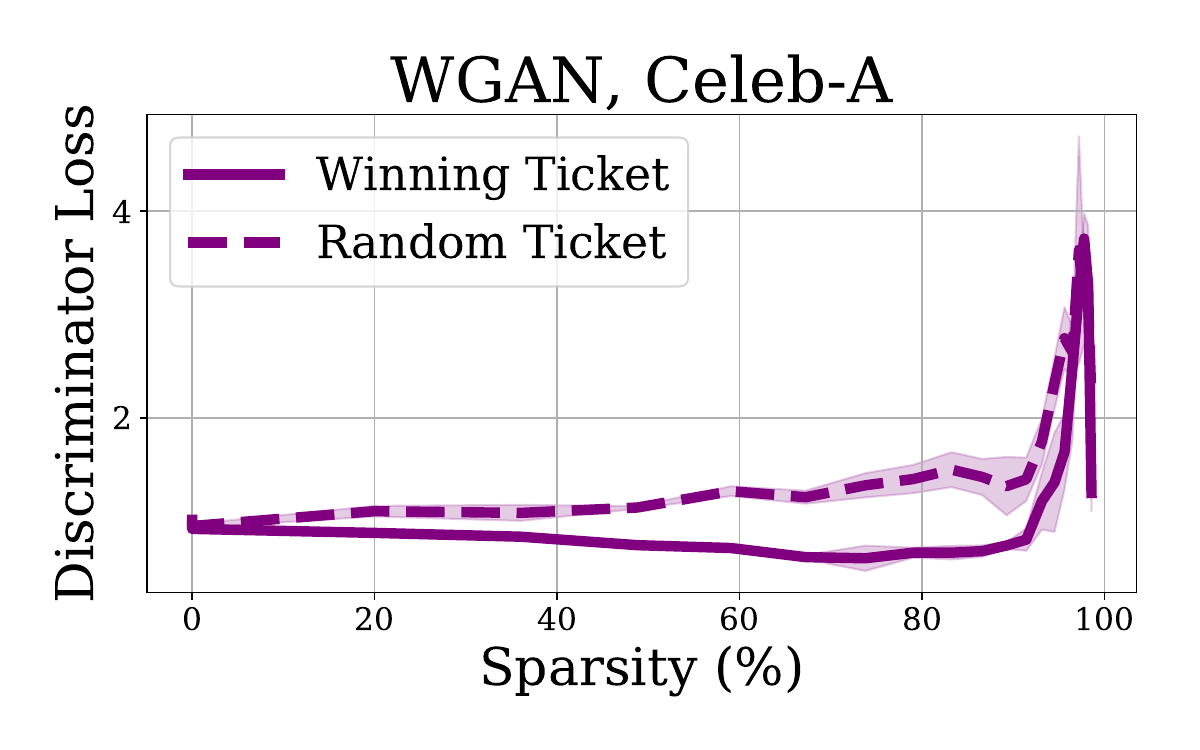}}

\caption{\textbf{Discriminator Loss. } We plot the discriminator loss of winning and random tickets of $4$ GAN models trained on CIFAR-10 and Celeb-A: DCGAN, ResNet-GAN, SNGAN and WGAN. Winning tickets consistently show lower loss. Experiments are performed on $5$ random runs, and the error bars represent $+/-$ standard deviation across runs.}
\label{gan_disc_plots}
\end{figure}

The discriminator in a GAN is responsible for distinguishing between fake and real images. In our experiments, apart from FID, we also use the discriminator loss as an evaluation metric to compare winning and random tickets. Figure \ref{gan_disc_plots} shows how random tickets lead to visibly higher discriminator loss than winning tickets across the $4$ types of GANs on both CIFAR-10 and Celeb-A. We also see in Figure \ref{gan_inception_plots}, that winning tickets have better inception scores than random tickets. We thus show that winning tickets exist and show good performance under various GAN setups when evaluated not only on the FID scores but also on the discriminator loss and inception scores.

\subsection{Images Generated by Winning and Random Tickets}
\label{images_sec}
\begin{figure*}[h]
\centering
\subfigure{\adjincludegraphics[width=\textwidth, trim={4.5cm 1.2cm 3.7cm 1.2cm}, clip]{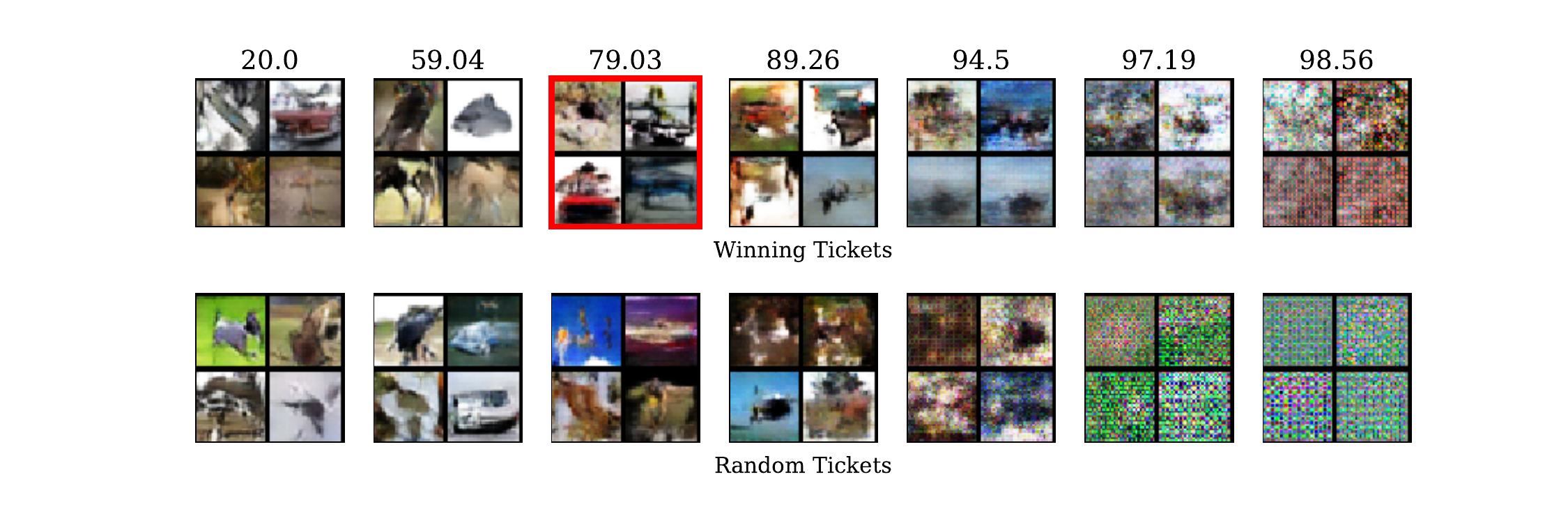}}
\subfigure{\adjincludegraphics[width=\textwidth, trim={4.5cm 1.2cm 3.7cm 1.2cm}, clip]{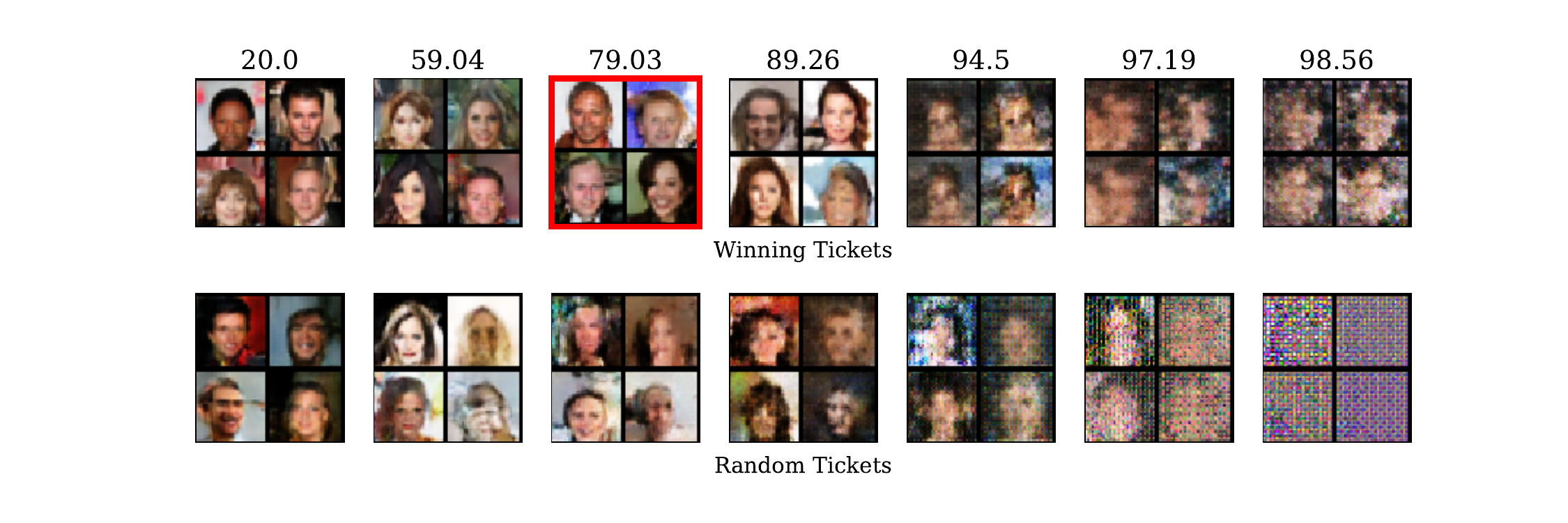}}
\caption{\textbf{DCGAN Generated Images. } Images generated by winning and random tickets of different sparsity levels on DCGAN on CIFAR-10 and Celeb-A. Winning tickets consistently show better quality images. The image highlighted in red corresponds to the winning ticket of highest sparsity.}
\label{gan_img}
\end{figure*}

\begin{figure*}[h]
\centering
\subfigure{\adjincludegraphics[width=\textwidth, trim={4.5cm 1.2cm 3.7cm 1.2cm}, clip]{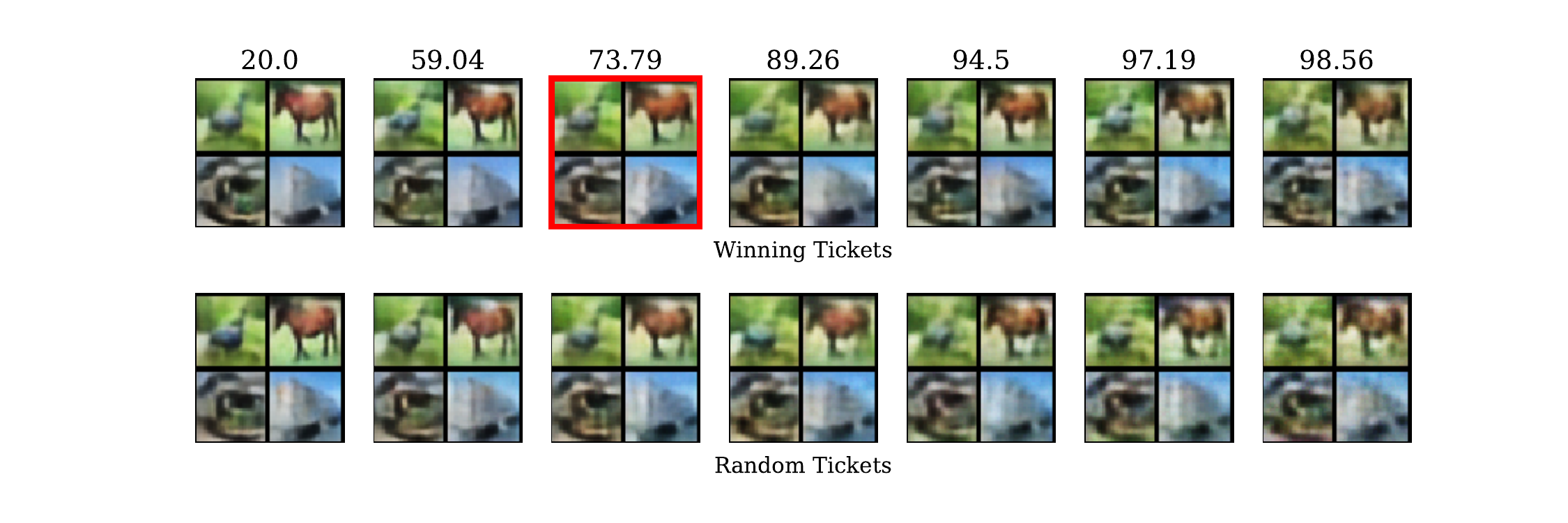}}
\subfigure{\adjincludegraphics[width=\textwidth, trim={4.5cm 1.2cm 3.7cm 1.2cm}, clip]{images/vae_Celeba.pdf}}
\caption{\textbf{VAE Generated Images. } Images generated by winning and random tickets of different sparsity levels on VAE on CIFAR-10 and Celeb-A. Winning tickets consistently show better quality images. The image highlighted in red corresponds to the winning ticket of highest sparsity.}
\label{vae_img}
\end{figure*}

\begin{figure*}[h]
\centering
\subfigure{\adjincludegraphics[width=\textwidth, trim={4.5cm 1.2cm 3.7cm 1.2cm}, clip]{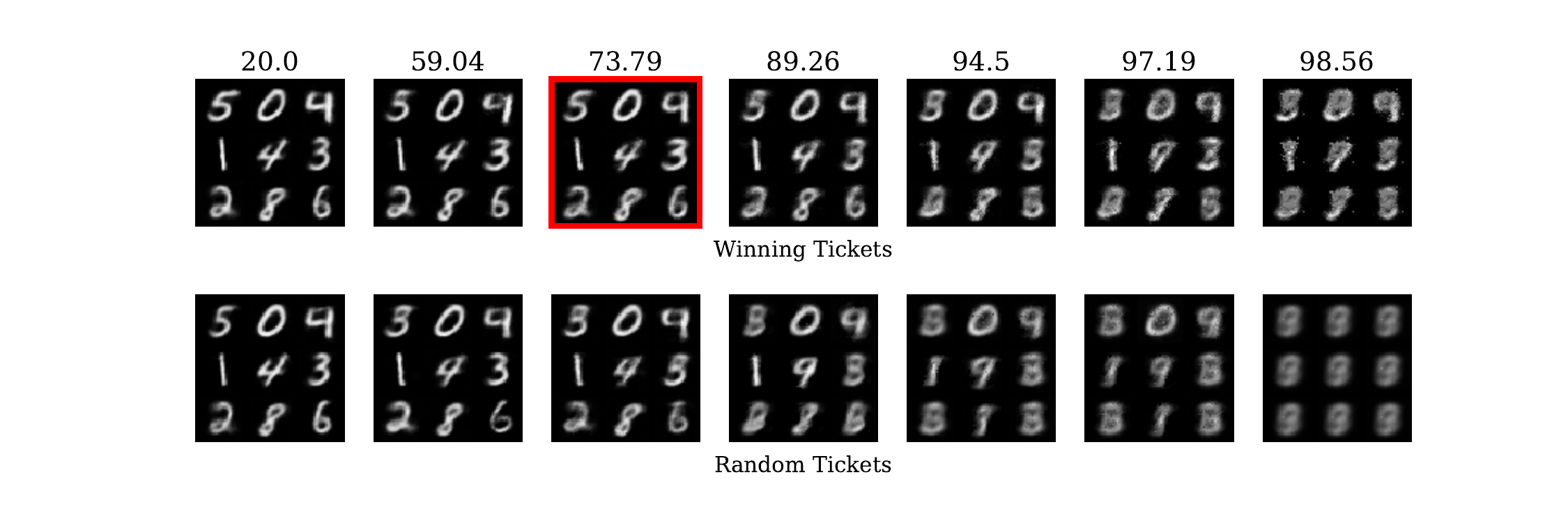}}
\subfigure{\adjincludegraphics[width=\textwidth, trim={4.5cm 1.2cm 3.7cm 1.2cm}, clip]{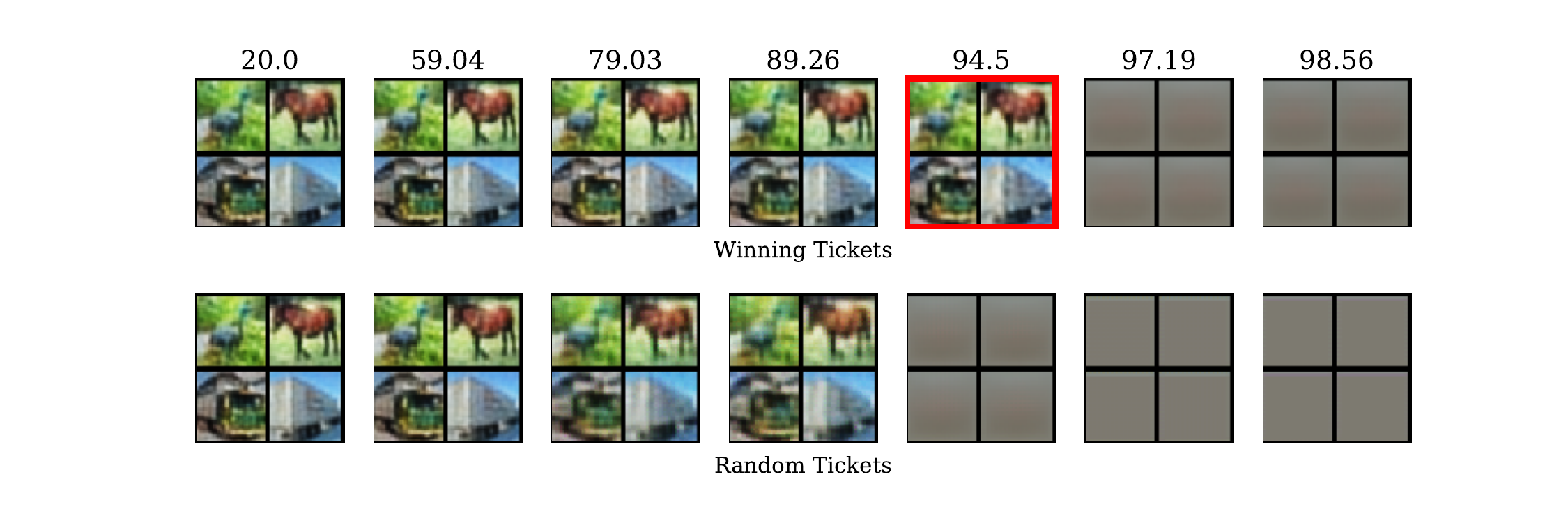}}
\subfigure{\adjincludegraphics[width=\textwidth, trim={4.5cm 1.2cm 3.7cm 1.2cm}, clip]{images/auto_Celeba.pdf}}
\caption{\textbf{AutoEncoder Generated Images. } Images generated by winning and random tickets of different sparsity levels on linear AutoEncoder on MNIST and convolutional AutoEncoder on CIFAR-10 and Celeb-A. Winning tickets consistently show better quality images. The image highlighted in red corresponds to the winning ticket of highest sparsity.}
\label{auto_img}
\end{figure*}

The quantitative evaluation measures used in the paper is helpful to show a comprehensive analysis and draw comparisons between winning and random tickets. However, for generative models, it is also important to validate winning tickets {\it qualitatively}. In AutoEncoder, as shown in Figure \ref{auto_img}, the reconstructed images maintain high quality until the pruning threshold (highlighted in red) of 73\% in MNIST, 94\% in CIFAR-10 and 98\% in Celeb-A which are the winning tickets of the highest sparsity. The corresponding random tickets reconstruct visibly poor quality images. In Figure \ref{vae_img}, we observe that VAE winning tickets maintain quality up to 73\% for CIFAR-10 and Celeb-A while random tickets lead to poorer image quality. In Figure \ref{gan_img}, for GANs, we see good winning ticket images up to 79\% on both CIFAR-10 and Celeb-A. Therefore, the winning tickets drawn from other evaluation metrics in this paper like FID, reconstruction loss and discriminator loss are validated by this qualitative analysis of generated images. 

\end{document}